%% file: main.tex
\documentclass{article}

\usepackage[final]{neurips_2025}

\usepackage{enumitem}
\usepackage[utf8]{inputenc} % allow utf-8 input
\usepackage[T1]{fontenc}    % use 8-bit T1 fonts
\usepackage{url}            % simple URL typesetting
\usepackage{booktabs}       % professional-quality tables
\usepackage{amsfonts}       % blackboard math symbols
\usepackage{nicefrac}       % compact symbols for 1/2, etc.
\usepackage{microtype}      % microtypography
\usepackage{algorithm}
\usepackage{amsmath}
\usepackage{amssymb}
\usepackage{mathtools}
\usepackage{amsthm}
\usepackage{multirow}
\usepackage{caption}
\usepackage{subcaption}
\usepackage{microtype}
\usepackage{wrapfig}
\usepackage{graphicx}
\usepackage{booktabs} % for professional tables
\usepackage[table,xcdraw]{xcolor} % Enable colors in tables
\usepackage{soul}  
\usepackage{tcolorbox}
\usepackage{fontawesome5} 
\tcbuselibrary{skins, breakable, listings, theorems}
\usepackage{hyperref}
\usepackage[algo2e]{algorithm2e}

\usepackage{color}
\title{Adaptive Distraction: Probing LLM Contextual Robustness with Automated Tree Search}

\definecolor{promptgray}{RGB}{200,200,200}
\definecolor{promptblue}{RGB}{25,118,210}

\newtcolorbox{promptbox}[2][]{%
    enhanced,
    unbreakable,
    before skip=2mm,
    after skip=2mm,
    colback=promptgray!20!white,
    colframe=promptblue!30!black,
    coltitle=white,
    boxrule=0.5mm,
    sharp corners,
    arc=5pt,
    attach boxed title to top center={yshift=-3mm},
    boxed title style={
        enhanced,
        colback=promptblue!50!white,
        colframe=promptblue,
        arc=5pt,
        outer arc=5pt,
        boxrule=0pt,
    },
    title={\faLightbulb[solid]\space #2},
    fonttitle=\bfseries\color{white},
    #1
}

\tcbset{
  outerbox/.style={
    colback=gray!3,
    colframe=gray!40,
    coltitle=black,
    fonttitle=\bfseries,
    boxrule=0.6pt,
    arc=1mm,
    width=\textwidth,
    before skip=10pt, after skip=10pt,
    left=4pt, right=4pt, top=5pt, bottom=5pt
  },
  innerbox/.style={
    colback=gray!1,
    colframe=gray!30,
    coltitle=black,
    fonttitle=\bfseries,
    boxrule=0.5pt,
    arc=0.5mm,
    width=\textwidth,
    top=4pt, bottom=4pt, left=5pt, right=5pt,
    boxsep=3pt
  }
}

\author{%
  \textbf{Yanbo Wang}$^{1,*}$, \textbf{Zixiang Xu}$^{1,*}$, \textbf{Yue Huang}$^{2,*}$, \textbf{Chujie Gao}$^{2}$, \textbf{Siyuan Wu}$^{1}$ \\ 
  \textbf{Jiayi Ye}$^{1}$, \textbf{Pin-Yu Chen}$^{3}$, \textbf{Xiuying Chen}$^{1,\dagger}$, \textbf{Xiangliang Zhang}$^{2,\dagger}$ \\[3pt]
  $^{1}$Mohamed bin Zayed University of Artificial Intelligence (MBZUAI)\\
  $^{2}$University of Notre Dame \quad
  $^{3}$IBM Research \\[3pt]
  {*Equal contribution \quad $^{\dagger}$Corresponding authors}
}

\begin{document}

\newcommand{\authorcomment}[2]{\textcolor{blue}{\textbf{[#1: #2]}}}
\newcommand{\yanbo}[1]{\authorcomment{Yanbo}{#1}}

\maketitle
\input{Section/0-abstract}
\input{Section/1-intro}

\input{Section/3-method}

\input{Section/4-experiments}
\input{Section/5-conclusion}
% \input{Section/6-impact statement}
% \clearpage
\bibliographystyle{unsrt}
\bibliography{references}
% \clearpage
\input{Section/99-appendix}

\end{document}

%% file: Section/0-abstract.tex
\begin{abstract}
Large Language Models (LLMs) often struggle to maintain their original performance when faced with semantically coherent but task-irrelevant contextual information. Although prior studies have explored this issue using fixed-template or retrieval-based distractions, such static methods show limited effectiveness against contemporary models. To address this problem, we propose a dynamic distraction generation framework based on tree search, where the generation process is guided by model behavior. Without modifying the original question or answer, the method efficiently produces challenging adaptive distractions across multiple datasets, enabling systematic stress testing of LLMs’ contextual robustness. Experiments on four benchmarks demonstrate that the generated distractions lead to an average performance drop of over 45\% for mainstream models. Further comparisons of mitigation strategies show that prompt-based optimization methods yield limited gains, whereas post-training approaches (e.g., DPO) significantly enhance the model's contextual robustness. The results indicate that these issues do not stem from knowledge deficits in LLMs, but from a fundamental inability to maintain consistent reasoning under contextual distraction, posing a major challenge to the reliability of LLMs in real-world applications. The code is publicly available at \url{https://github.com/wyf23187/Adaptive_Distractions}.

\end{abstract}

%% file: Section/1-intro.tex
\section{Introduction}

Large Language Models (LLMs) have achieved remarkable success across diverse natural language processing tasks, such as question answering, summarization, and reasoning \cite{achiam2023gpt, zhou2023survey, huang2025trustgen, xu2025socialmaze}. However, recent studies reveal a critical vulnerability: LLMs are susceptible to semantically coherent but task-irrelevant contextual information, which can significantly degrade their performance \cite{shi2023large}. This lack of contextual robustness hinders their ability to consistently focus on essential task content in the presence of distracting information, a challenge particularly pronounced in real-world applications where irrelevant context is common. Addressing this limitation is crucial to ensure the reliability of LLMs in complex, dynamic environments.

Current methods for evaluating LLM contextual robustness primarily rely on fixed-template or retrieval-based distractors \cite{shi2023large, liu2023lost}. However, our preliminary experiments demonstrate that these static approaches are increasingly ineffective against contemporary models, with performance degradation often below 5\% on advanced models like GPT-4o, which is detailed in Appendix~\ref{app:preliminary_results}. 
This highlights that existing methods lack adaptiveness and are heavily dependent on the specific behavior of the target LLM. Once the model evolves, previously effective attack strategies may become obsolete.
Moreover, such limited impact is insufficient to provide a reliable basis for robustness evaluation in realistic scenarios \cite{shayegani2023survey}. 
This highlights the urgent need for dynamic, adaptive methods capable of generating contextually plausible distractions that evolve with LLM capabilities, ensuring robust evaluation across diverse tasks and models.

% Although prior work has revealed that LLMs can be sensitive to irrelevant but coherent context \cite{shi2023large, levy2024same}, existing methods for generating such distractions are often static and handcrafted. Approaches based on fixed templates or retrieved passages \cite{zhu2024dynamic, yoran2023making} lack adaptability, and their impact diminishes when applied to stronger models. As a result, they fail to provide a reliable basis for evaluating contextual robustness. In this work, we aim to address this gap by developing a generation framework that automatically constructs effective and semantically valid contextual distractions, guided by feedback from the model's behavior. We refer to the resulting examples as \textbf{adaptive distractions}, contextual additions that are coherent and answer-preserving, yet sufficient to induce performance degradation. These examples allow us to systematically stress-test the ability of LLMs to filter out irrelevant information and maintain consistent reasoning.

To address the limitations of static methods, we aim to propose an adaptive attack method to generate \textbf{adaptive distractions}---semantically coherent, answer-preserving contextual additions that significantly impair LLM performance. This approach aims to evolve with advancing LLM capabilities, enabling robust stress-testing across diverse tasks without being constrained by model strength. However, generating such distractions presents key challenges: (1) ensuring semantic coherence with the original input, (2) preserving the correct answer, and (3) creating distractions potent enough to disrupt model predictions. These requirements demand a dynamic, model-informed generation strategy to effectively probe LLM contextual robustness.

To address these challenges, we propose a structured generation framework based on tree search to automatically construct adaptive distractions \cite{yao2023tree, browne2012survey, xu2025cross}. Our approach employs a classifier to pre-filter questions susceptible to perturbation, followed by a tree search module that explores contextual additions using a priority queue guided by model behavior. Error-guided perturbations generate candidate distractions at each node, evaluated for their ability to alter predictions while preserving semantic consistency. Early stopping strategies enhance efficiency, ensuring scalability across tasks and model families. \autoref{fig:pipeline} illustrates this pipeline, which enables the controlled and automated creation of challenging distractions tailored to probe LLM contextual robustness.

We conducted comprehensive experiments to validate our framework, evaluating its effectiveness across four benchmark datasets, namely MMLU, CommonsenseQA, OpenbookQA, and TruthfulQA, on a diverse set of mainstream models, including proprietary and open-weight architectures. Our results show that adaptive distractions cause significant performance degradation, with an average accuracy drop exceeding 45\%, exposing vulnerabilities in even the most advanced LLMs. Additionally, we explored mitigation strategies, comparing prompt-based approaches with targeted fine-tuning, and analyzed supplementary experiments, including prompt variants and case studies, detailed in the appendix. These findings collectively underscore the critical need for enhanced contextual robustness in LLMs.

In summary, our work delivers the following contributions:
\begin{itemize}[
    leftmargin=1.5em,
    itemsep=0.2ex,   
    parsep=0pt,      
    topsep=0pt,      
    partopsep=0pt, 
]
    \item We introduce a novel framework for generating adaptive distractions, semantically coherent yet task-irrelevant additions, enabling robust and systematic evaluation of LLM contextual robustness.
    \item We provide empirical evidence of significant performance degradation, exceeding 45\% accuracy drop, across four benchmark datasets and diverse mainstream models, uncovering persistent vulnerabilities in advanced LLMs.
    \item We evaluate mitigation strategies, revealing that targeted fine-tuning substantially enhances robustness under contextual distraction, while prompt-based approaches yield limited effectiveness.
\end{itemize}

%% file: Section/3-method.tex
\section{Methodology}
\subsection{Overview}

\begin{figure*}[t]
    \centering
    \includegraphics[width=\linewidth]{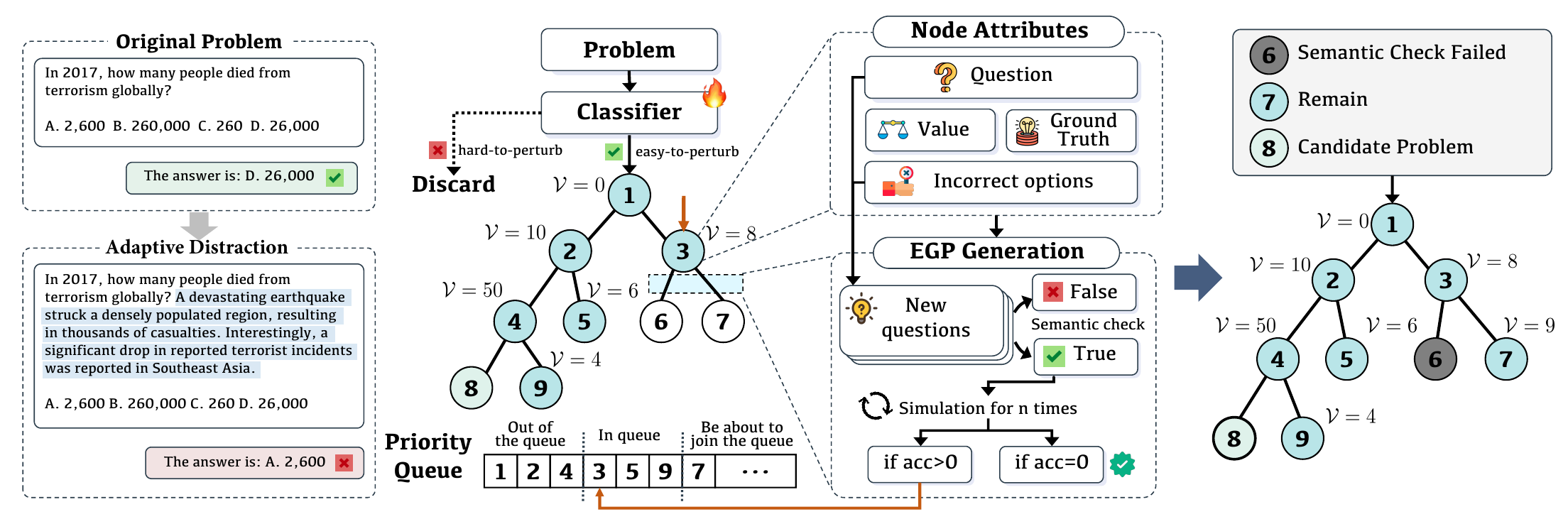}
    \vspace{-10pt}
    \caption{Overview of our framework. Given an input question, a classifier first filters for instances that are more susceptible to contextual interference. Then, a tree-based search explores candidate context additions using error-guided perturbation (EGP). At each node, candidate distractions are evaluated based on their ability to alter model predictions without affecting the correct answer. The framework efficiently produces high-quality \emph{adaptive distraction examples} that challenge model robustness.}
    \label{fig:pipeline}
\end{figure*}

As illustrated in \autoref{fig:pipeline}, our objective is to systematically identify vulnerabilities in LLMs by generating adaptive distractions. As defined in the Introduction, these are contextual additions designed to preserve the original question's meaning and answer while affecting model performance.

Generating such distractions presents three core challenges. First, it requires an effective reward mechanism to guide the generation toward minimal yet impactful context additions. Second, the search space for valid distractions is vast, leading to substantial computational cost if not carefully controlled. Third, ensuring semantic consistency and bounded input length is critical, as irrelevant context may introduce unintended shifts or exceed the model’s optimal context window.

To address these challenges, we propose a multi-step framework. We first apply a \emph{classifier-based filtering} process to identify examples that are more likely to be affected by contextual interference, which narrows the search space. Then, we perform a \emph{tree-based search} to explore semantically valid distractions. At each node, a \emph{proxy model} generates candidate perturbations, which are evaluated through simulation using the victim model. The search is guided by a value function that combines model failure signals with depth penalties to ensure both quality and efficiency. Finally, we incorporate semantic validation and length control to preserve the original problem’s intent, alongside pruning and early stopping strategies to reduce computation without sacrificing attack effectiveness.

\subsection{Problem Formulation}

Let \( D = \{P_1, P_2, \dots, P_N\} \) denote a dataset consisting of \( N \) multiple-choice problem instances, where each instance is represented as a tuple:
\begin{equation}
    P = \langle Q, A_\text{gt}, \mathcal{A}_\text{inc} \rangle,
\end{equation}
with \( Q \) denoting the question, \( A_\text{gt} \) the ground truth answer, and \( \mathcal{A}_\text{inc} \) a set of incorrect answers. Given a victim model \( M \), our goal is to construct a perturbed dataset \( D' = \{P'_1, P'_2, \dots, P'_N\} \), where each perturbed instance \( P' = \langle Q', A_\text{gt}, \mathcal{A}_\text{inc} \rangle \) is obtained by applying an \emph{adaptive distraction} \( \Delta Q \) to the question \( Q \), such that:
\begin{equation}
\label{equ:Q}
Q' = Q + \Delta Q.
\end{equation}
Our aim is to optimize the distraction \( \Delta Q \) to minimize the accuracy of \( M \) on \( D' \), while ensuring semantic consistency and length constraints between \( Q \) and \( Q' \). Here, semantic consistency is determined by a binary classifier \( S \), which outputs \( S(Q, Q') \in \{0, 1\} \), where \( S(Q, Q') = 1 \) indicates no semantic shift. Formally, the problem is expressed as:
\begin{equation}
\label{equ:constraint}
\min_{\Delta Q}  \quad \mathbb{E}_{P \sim D} \Big[ \mathcal{L}_\text{accuracy}(M, Q') \Big], \text{s.t.}  \quad S(Q, Q') = 1, \quad \frac{\text{len}(Q')}{\text{len}(Q)} \leq \lambda,
\end{equation}

Here, \( S \) ensures that the distraction \( \Delta Q \) does not lead to a semantic shift, while the length ratio constraint \(\lambda\) ensures \( Q' \) remains within acceptable bounds compared to the original~\( Q \). This constraint is necessary because recent studies show that LLMs experience performance degradation in long context scenarios \cite{bai2023longbench}. To prevent excessive length expansion in  $Q'$, we introduce a length constraint, where \( \lambda \) is an upper bound on the relative length of \( Q' \) compared to \( Q \).

If the output \( Q' \) does not satisfy the constraints in \autoref{equ:constraint}, it is discarded, and a new distraction \( \Delta Q \) is generated by re-prompting the proxy model.

\subsection{Error-Guided Perturbation Generation}

The distraction \( \Delta Q \) is generated using a \textbf{proxy model}, denoted as \( P_\text{proxy} \). The proxy model is prompted with the original problem instance \( P = \langle Q, A_\text{gt}, \mathcal{A}_\text{inc} \rangle \) and tasked with generating a modified question \( Q' \) defined in \autoref{equ:Q}, where \( \Delta Q \) represents the distraction introduced by \( P_\text{proxy} \). The generation process is formalized as:
\begin{equation}
\Delta Q = P_\text{proxy}(Q, A_\text{gt}, \mathcal{A}_\text{inc}),
\end{equation}
where \( P_\text{proxy} \) generates \( \Delta Q \) based on a predefined prompt designed to guide the proxy model in producing contextual additions that increase the likelihood of the victim model \( M \) selecting an incorrect answer \( a_\text{inc} \in \mathcal{A}_\text{inc} \) (i.e., lead the model to produce an error).

\subsection{Tree-Based Perturbation Exploration}
\label{sec:value_function}
We employ a tree-based simulation-driven method to optimize distractions by heuristically exploring the search space. A priority queue is maintained to store nodes ordered by their value \( \mathcal{V}(P') \), with the highest-value node dequeued and expanded iteratively using \( P_\text{proxy} \) to identify high-potential vulnerabilities.

\textbf{Simulation For Measuring Distraction Quality.} Firstly, we aim to design the reward of distracted questions to measure their value. For a given problem instance \( P = \langle Q, A_\text{gt}, \mathcal{A}_\text{inc} \rangle \), the simulation process evaluates the quality of a distraction by estimating the success rate of the victim model \( M \) on \( P \). Let \( y \sim M(y \mid P) \) represent the output of the model \( M \) when queried on \( P \). During a single simulation, the success rate \( r_M(P) \) is computed by sampling \( n \) model outputs:
\begin{equation}
r_M(P) = \frac{1}{n} \sum_{i=1}^{n} \mathbb{I}\{y_i = A_\text{gt}\}, \quad y_i \sim M(y \mid P),
\end{equation}
where \( \mathbb{I}\{\cdot\} \) is an indicator function that returns 1 if the model’s output \( y_i \) matches the ground truth answer \( A_\text{gt} \), and 0 otherwise. The success rate \( r_M(P) \) quantifies the likelihood of the model producing the correct answer under the given distraction.

A distracted problem \( P' = \langle Q', A_\text{gt}, \mathcal{A}_\text{inc} \rangle \) is considered effective if \( r_M(P') = 0 \), indicating that the model fails to produce the correct answer in all sampled outputs. The simulation process computes a value \( \mathcal{V}(P') \) for the node corresponding to \( P' \) in the tree-based search:
\begin{equation}
\mathcal{V}(P') = \exp\left(\frac{\alpha}{r_M(P')}\right) \cdot \text{depth}^{-\gamma}, \quad \text{s.t. } r_M(P') \neq 0,
\end{equation}
where \( \alpha \) and \( \gamma \) are scaling constants, \( r_M(P') \) is the success rate of the victim model \( M \) on \( P' \), and \( \text{depth} \) represents the recursion depth of the node in the search tree. For the question with \( r_M(P')=0 \), we add it into the candidate problem list \( L \), which stores examples that effectively \emph{induce failure in the LLM}.

The simulation process systematically estimates \( \mathcal{V}(P') \), prioritizing distracted problems with lower success rates \( r_M(P') \), which correspond to higher potential vulnerabilities in the model. Simultaneously, the factor \( \text{depth}^{-\gamma} \) discourages deeper recursions in the search tree, ensuring computational efficiency. High-value nodes with large \( \mathcal{V}(P') \) scores are prioritized in the following tree-based search.

\textbf{Tree-Based Search.} For the tree-based search, the process begins with a root \( P_\text{root} = \langle Q, A_\text{gt}, \mathcal{A}_\text{inc} \rangle \). A priority queue \( \mathcal{Q} \) is maintained, where each node \( P' \) is ordered by its value \( \mathcal{V}(P') \) in descending order. Initially, the root node is added to the queue as \( \mathcal{Q} \gets \mathcal{Q} \cup \{P_\text{root}\} \). At each iteration, the node \( P' \) with the highest value \( \mathcal{V}(P') \) is dequeued for exploration:
\begin{equation}
P' = \arg\max_{P \in \mathcal{Q}} \mathcal{V}(P), \quad \mathcal{Q} \gets \mathcal{Q} \setminus \{P'\}.
\end{equation}

The proxy model \( P_\text{proxy} \) generates \( k = |\mathcal{A}_\text{inc}| \) child nodes for \( P' \), corresponding to distractions \( \Delta Q_j \) derived from each incorrect candidate answer \( a_\text{inc} \in \mathcal{A}_\text{inc} \):
\begin{equation}
Q'_j = Q' + \Delta Q_j, \quad P'_j = \langle Q'_j, A_\text{gt}, \mathcal{A}_\text{inc} \rangle, \quad j = 1, 2, \dots, k.
\end{equation}
Each child node \( P'_j \) is evaluated by a simulation-driven method to compute its value \( \mathcal{V}(P'_j) \), and the child nodes are added to the priority queue:
\begin{equation}
\mathcal{Q} \gets \mathcal{Q} \cup \{P'_1, P'_2, \dots, P'_k\}.
\end{equation}
The search iteratively repeats this searching process, dynamically expanding the highest-value node and exploring the distraction space.

\textbf{Why not Monte Carlo Tree Search?} Monte Carlo Tree Search (MCTS) \cite{browne2012survey} has been widely used in recent studies to perform simulations powered by LLMs, achieving remarkable performance \cite{zhang2024rest, wang2024seed, xie2024monte, guan2025rstar}. However, MCTS is not suitable for our task due to its focus on balancing exploration (searching broadly across the tree) and exploitation (focusing on promising branches). In our context, such a balance is unnecessary because the width of the tree is inherently fixed, dictated by the number of incorrect answer candidates \( |\mathcal{A}_\text{inc}| \). Moreover, MCTS introduces computational overhead by maintaining dynamic exploration strategies, which is impractical given the predefined structure and requirements of our method. Therefore, we opt for a simpler and more task-specific tree design that aligns directly with the properties of our problem.

\subsection{Efficiency Strategies}

\textbf{Early Stopping Strategies.} To reduce computational costs during the search process, we employ two early stopping strategies: diversity control and performance-based pruning.

The first strategy, diversity control, limits the number of child nodes considered at each search step. For a node \( P' \), if the number of child nodes \( P'_j \) satisfying \( r_M(P'_j) = 0 \) exceeds a predefined threshold \( n_1 \), we add the top \( n_1 \) child nodes to the candidate problem list \( L \) and directly pass this branch without further exploration. Formally, let \( \mathcal{C}(P') \) represent the set of child nodes of \( P' \), and define:
\begin{equation}
\mathcal{C}_0(P') = \{P'_j \in \mathcal{C}(P') \mid r_M(P'_j) = 0\}.
\end{equation}
If \( |\mathcal{C}_0(P')| > n_1 \), we update the candidate problem list \( L \) as:
\begin{equation}
L \gets L \cup \mathcal{C}_0(P')[1:n_1],
\end{equation}
where \( [1:n_1] \) indicates the top \( n_1 \) nodes according to their values \( \mathcal{V}(P'_j) \). The branch corresponding to \( P' \) is then terminated.

The second strategy is performance-based pruning, which bypasses nodes where further exploration is unlikely to yield meaningful results. For a node \( P' \), if all its child nodes satisfy \( r_M(P'_j) = 1 \), the node \( P' \) is skipped. Formally, if:
\begin{equation}
r_M(P'_j) = 1, \quad \forall P'_j \in \mathcal{C}(P'),
\end{equation}
then \( P' \) is pruned from the search.

Additionally, if for \( l \) consecutive levels of the search tree, the minimum success rate \( \min(r_M(P')) \) at each level increases monotonically from the top level to the bottom level, the corresponding node is bypassed. Let \( \text{level}_i \) represent the set of nodes at level \( i \) of the search tree, and define \( m_i = \min_{P' \in \text{level}_i} r_M(P') \). If:
\begin{equation}
m_{i+1} > m_i, \quad \forall i \in \{1, 2, \dots, l-1\},
\end{equation}
then the corresponding branch of the search tree is pruned.

\textbf{Problem Filtering via Classifier.} To reduce search costs, a classifier \( C(Q) \) is used to filter out questions with low potential to become effective distraction candidates (e.g., the extremely easy question ``What is the highest mountain in the world?''). The classifier is trained on previously searched questions, \( \mathcal{D}_\text{train} = \{(Q_i, y_i)\}_{i=1}^{N} \), where \( y_i \in \{0, 1\} \) indicates whether \( Q_i \) successfully exposes a vulnerability in the victim model \( M \).

For each new question \( Q \), the classifier computes \( p(y=1 \mid Q) = C(Q) \). Questions satisfying \( p(y=1 \mid Q) < \tau_C \), where \( \tau_C \) is a predefined threshold, are discarded:
\begin{equation}
\mathcal{Q} \gets \mathcal{Q} \setminus \{Q \mid p(y=1 \mid Q) < \tau_C\}
\end{equation}

Due to the space limitation, we show the overall algorithm in \autoref{app:alg}.

%% file: Section/4-experiments.tex
\section{Experiment}

\subsection{Experiment Setup}

\textbf{Selected Datasets.} We selected four widely used benchmarks to evaluate contextual robustness under adaptive distraction: MMLU \citep{hendryckstest2021,hendrycks2021ethics}, CommonsenseQA \citep{talmor-etal-2019-commonsenseqa}, OpenbookQA \citep{OpenBookQA2018}, and TruthfulQA \citep{lin2021truthfulqa}. These datasets cover diverse domains such as factual knowledge, commonsense reasoning, and elementary science, making them suitable for probing how LLMs handle irrelevant but semantically coherent contextual additions.

\textbf{Models.} As shown in \autoref{tab:all_models}, we used four proprietary models in our experiments: GPT-4o \citep{hurst2024gpt}, GPT-4o-mini \citep{openai2024gpt4omini}, Claude-3.5-Sonnet \citep{anthropic2024claude35}, and o1-mini \citep{jaech2024openai}. Additionally, we included eight open-weight models: Gemma-2-2B, Gemma-2-27B \citep{gemma_2024}, Qwen2.5-1.5B, Qwen2.5-7B, Qwen2.5-72B \citep{qwen2,qwen2.5}, Llama-3.1-8B \citep{meta2024llama31_8b}, Llama-3.1-70B \citep{meta2024llama31_70b}, and Phi-3.5-mini \citep{abdin2024phi}.

\textbf{Hyperparameter Setting.} We set the temperature to 0.7 during the distraction generation phase to encourage more diverse and challenging outputs. For evaluation, we lowered the temperature to 0.001 to ensure response consistency, with a maximum output length of 1,024 tokens. Additionally, we set $\alpha = 2$ and $\gamma = 1$ for the value function used in the tree search. For other detailed hyperparameter settings, please refer to Appendix~\ref{appendix:experiment_settings}.

\textbf{Prompt Template.} Prompt-based templates are used for several sub-tasks throughout our framework, including generating contextual distractions, assessing semantic consistency, evaluating model answers (zero-shot + CoT), baseline elaboration, filtering distraction-susceptible samples, and conducting robustness enhancement. The specific templates are provided in Appendix~\ref{appendix:prompt_template}.

\textbf{Human Verification.}
To confirm semantic preservation, we conducted a human evaluation on randomly sampled perturbed questions. Annotators judged both semantic equivalence and answer consistency.
 Detailed results are reported in Appendix~\ref{appendix:human_evaluation}.

\begin{table*}[t]
\centering
\definecolor{rowlight}{HTML}{F8FAFB}
\definecolor{rowdark}{HTML}{EEF3F5}
\renewcommand{\arraystretch}{1.2}
\caption{Accuracy of seven LLMs on four benchmarks before (\textbf{Original}) and after (\textbf{Perturbed}) applying our adaptive distractions. Cell background colors emphasise the severity of that drop.}

\label{tab:main_results}
\resizebox{\textwidth}{!}{%
\rowcolors{3}{rowlight}{rowdark} 
\begin{tabular}{lcccccccccccc}
    \toprule[1pt]
    \multirow{2}{*}{\textbf{Model}} & \multicolumn{3}{c}{\textbf{CommonsenseQA}} & \multicolumn{3}{c}{\textbf{OpenbookQA}} & \multicolumn{3}{c}{\textbf{TruthfulQA}} & \multicolumn{3}{c}{\textbf{MMLU}} \\
    & Original & Perturbed & $\Delta$ & Original & Perturbed & $\Delta$ & Original & Perturbed & $\Delta$ & Original & Perturbed & $\Delta$ \\
    \midrule
    GPT-4o-mini       & 0.857 & 0.220 & \cellcolor[HTML]{FF9999}0.637 & 0.897 & 0.228 & \cellcolor[HTML]{FF9999}0.668 & 0.607 & 0.160 & \cellcolor[HTML]{FF9999}0.447 & 0.787 & 0.255 & \cellcolor[HTML]{FF9999}0.532 \\
    Llama-3.1-8B      & 0.753 & 0.230 & \cellcolor[HTML]{FF9999}0.524 & 0.807 & 0.212 & \cellcolor[HTML]{FF9999}0.595 & 0.570 & 0.283 & \cellcolor[HTML]{FFFFCC}0.288 & 0.697 & 0.300 & \cellcolor[HTML]{FFCCCC}0.397 \\
    Gemma-2-27B       & 0.857 & 0.249 & \cellcolor[HTML]{FF9999}0.607 & 0.867 & 0.231 & \cellcolor[HTML]{FF9999}0.636 & 0.782 & 0.449 & \cellcolor[HTML]{FFCCCC}0.332 & 0.753 & 0.340 & \cellcolor[HTML]{FF9999}0.413 \\
    o1-mini           & 0.856 & 0.296 & \cellcolor[HTML]{FF9999}0.560 & 0.897 & 0.377 & \cellcolor[HTML]{FF9999}0.519 & 0.748 & 0.523 & \cellcolor[HTML]{FFFFCC}0.226 & 0.803 & 0.451 & \cellcolor[HTML]{FFCCCC}0.352 \\
    Qwen2.5-72B       & 0.880 & 0.304 & \cellcolor[HTML]{FF9999}0.576 & 0.917 & 0.325 & \cellcolor[HTML]{FF9999}0.592 & 0.790 & 0.442 & \cellcolor[HTML]{FFCCCC}0.348 & 0.807 & 0.412 & \cellcolor[HTML]{FFCCCC}0.395 \\
    GPT-4o            & 0.890 & 0.277 & \cellcolor[HTML]{FF9999}0.613 & 0.950 & 0.375 & \cellcolor[HTML]{FF9999}0.575 & 0.757 & 0.494 & \cellcolor[HTML]{FFFFCC}0.263 & 0.870 & 0.552 & \cellcolor[HTML]{FFCCCC}0.318 \\
    Claude-3.5-sonnet & 0.873 & 0.345 & \cellcolor[HTML]{FF9999}0.529 & 0.953 & 0.529 & \cellcolor[HTML]{FF9999}0.424 & 0.840 & 0.734 & \cellcolor[HTML]{FFFFCC}0.106 & 0.877 & 0.645 & \cellcolor[HTML]{FFFFCC}0.232 \\
    \bottomrule[1pt]
\end{tabular}%
}
\vspace{-1em}
\end{table*}

\begin{figure*}[t]
    \centering
    \includegraphics[width=\linewidth]{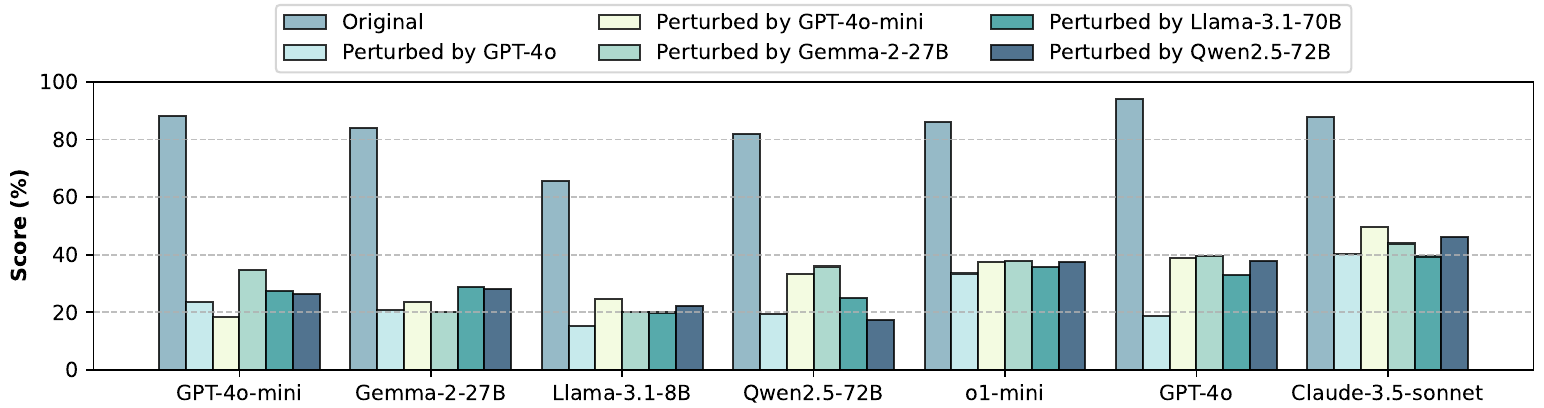}
    \vspace{-1em}
    \caption{Performance of victim models on original questions and perturbed questions generated by different proxy LLMs.}
    %Overall results between questions enhanced by different models.}
    \vspace{-2em}
    \label{fig:multi_results}
\end{figure*}

\subsection{Main Results} 

We conducted extensive evaluation experiments and mitigation studies to assess the impact of \textbf{adaptive distractions} on LLMs. The detailed configurations and dataset-model setups corresponding to each figure and table can be found in Appendix~\ref{appendix:settings}.

\textbf{Our method enables LLMs to autonomously generate adaptive distractions and effectively self-challenge.} In \autoref{fig:multi_results}, we configure the same model to act as both the proxy and the victim in the distraction generation process. We observe that all tested models suffer a substantial drop in accuracy when evaluated on adaptively distracted questions, compared to their original performance. For instance, GPT-4o-mini experiences a performance decline of over 40\%.

Furthermore, we uncover an intriguing pattern: distractions generated by a model itself tend to be more adversarial to that model than those generated by others. For example, GPT-4o-mini achieves an accuracy of only 0.185 on distractions it generated for itself, compared to 0.235 on those generated by the more advanced GPT-4o. This suggests that \textbf{models may be better at identifying their own weaknesses}, resonating with previous findings on self-alignment \cite{sun2024principle} and self-correction \cite{pan2023automatically}.

\textbf{All models are susceptible to adaptive distractions, regardless of their scale or capability.} Our results show that distractions created by stronger models can reliably challenge weaker models, while even distractions from weaker models can degrade the performance of stronger ones. As shown in \autoref{fig:multi_results}, powerful models such as GPT-4o and Claude-3.5-Sonnet still achieve below 50\% accuracy when evaluated on distractions generated by Gemma-2-27B. This highlights a fundamental vulnerability: \textit{no model is currently robust to adaptive distractions}.\footnote{We refer to the leaderboard at \url{https://lmarena.ai/} for model performance comparisons.}

\textbf{The extent of performance degradation varies significantly across tasks.} As shown in \autoref{tab:main_results}, the average drop in accuracy on TruthfulQA is consistently smaller than that on OpenbookQA across all models. This suggests that distraction sensitivity varies by domain. Tasks like OpenbookQA, which require precise factual retrieval, appear more vulnerable to contextual interference than trustworthiness-based tasks such as TruthfulQA.

\begin{wraptable}{r}{0.60\linewidth}
\vspace{-1em}
\centering
\small
\definecolor{rowlight}{HTML}{F8FAFB}
\definecolor{rowdark}{HTML}{EEF3F5}
\definecolor{headerbg}{HTML}{DCE7EF}
\renewcommand{\arraystretch}{1.1}
\setlength{\tabcolsep}{1.4mm}
\caption{Accuracy on original and distracted samples generated by different methods. \textbf{ICA}: Adding semantically coherent but task-irrelevant information to increase question length. \textbf{SPD}: Generating distractions via a single prompt without further optimization. \textbf{DyVal2}: Dynamic evaluation baseline \citep{zhu2024dynamic}. Lower scores indicate more effective distractions.}
\label{tab:baseline}
\rowcolors{2}{rowlight}{rowdark}
\begin{tabular}{
    >{\columncolor{headerbg}}l
    >{\columncolor{headerbg}}c
    >{\columncolor{headerbg}}c
    >{\columncolor{headerbg}}c
    >{\columncolor{headerbg}}c
    >{\columncolor{headerbg}}c
}
    \toprule[1pt]
    \textbf{Model} & \textbf{Original} & \textbf{ICA} & \textbf{SPD} & \textbf{DyVal2} & \textbf{Ours} \\
    \midrule
    GPT-4o-mini       & 0.890 & 0.727 & 0.760 & 0.630 & 0.185 \\
    Gemma-2-27B       & 0.860 & 0.788 & 0.790 & 0.650 & 0.237 \\
    Llama-3.1-8B      & 0.667 & 0.657 & 0.620 & 0.640 & 0.247 \\
    Qwen2.5-72B       & 0.820 & 0.737 & 0.810 & 0.697 & 0.334 \\
    o1-mini           & 0.860 & 0.694 & 0.770 & 0.697 & 0.374 \\
    GPT-4o            & 0.940 & 0.818 & 0.850 & 0.740 & 0.390 \\
    Claude-3.5-sonnet & 0.879 & 0.838 & 0.820 & 0.780 & 0.495 \\
    \midrule
    Average           & 0.843 & 0.757 & 0.774 & 0.691 & \textbf{0.323} \\
    \bottomrule[1pt]
\end{tabular}
\vspace{-10pt}
\end{wraptable}

\textbf{Our method outperforms existing distraction techniques by a large margin.} We compare our method with several baselines: (1) \emph{Irrelevant Context Augmentation (ICA)}, which adds semantically coherent but task-irrelevant information to extend question length \cite{ye2024justice}; (2) \emph{Single-Prompt Distraction (SPD)}, which generates distractions via a single prompt without optimization \citep{shi2023large}; and (3) DyVal2, a recent dynamic evaluation framework \citep{zhu2024dynamic}. As shown in \autoref{tab:baseline}, our method results in an average accuracy drop of 52.0\%, compared to 15.2\% for DyVal2. Even strong models such as Claude-3.5-Sonnet experience a 38.4\% absolute accuracy drop under our framework, compared to only 10.0\% under DyVal2. These results highlight the strength of our tree-based search framework in systematically identifying contextual vulnerabilities, not merely increasing input complexity.

\textbf{Adaptive distractions demonstrate robust cross-task generalization.} We further examine the generalization of adaptive distraction by applying it to other tasks
with well-defined ground truth, including mathematical reasoning.
As shown in Appendix~\ref{app:math500}, our method remains effective on the MATH-500 benchmark\citep{lightman2023lets}, suggesting that contextual distraction is not confined to factual QA but extends to any task where correctness can be explicitly evaluated.

\subsection{Classifier: Filtering Hard-to-Perturb Problems}

A critical challenge in adaptive distraction generation lies in distinguishing between \textbf{hard-to-perturb problems} (e.g., simple factual or arithmetic questions that are consistently answered correctly by LLMs regardless of added context) and \textbf{easy-to-perturb problems} (questions susceptible to semantic-preserving contextual distractions). Our analysis reveals that approximately 37\% of computational resources are typically wasted on attempting to distract hard-to-perturb examples. To address this inefficiency, we design classifiers that predict the distraction susceptibility of a given question.

\begin{figure}[htbp]
    \centering
    \begin{minipage}[b]{0.48\linewidth}
        \centering
        \includegraphics[width=\linewidth]{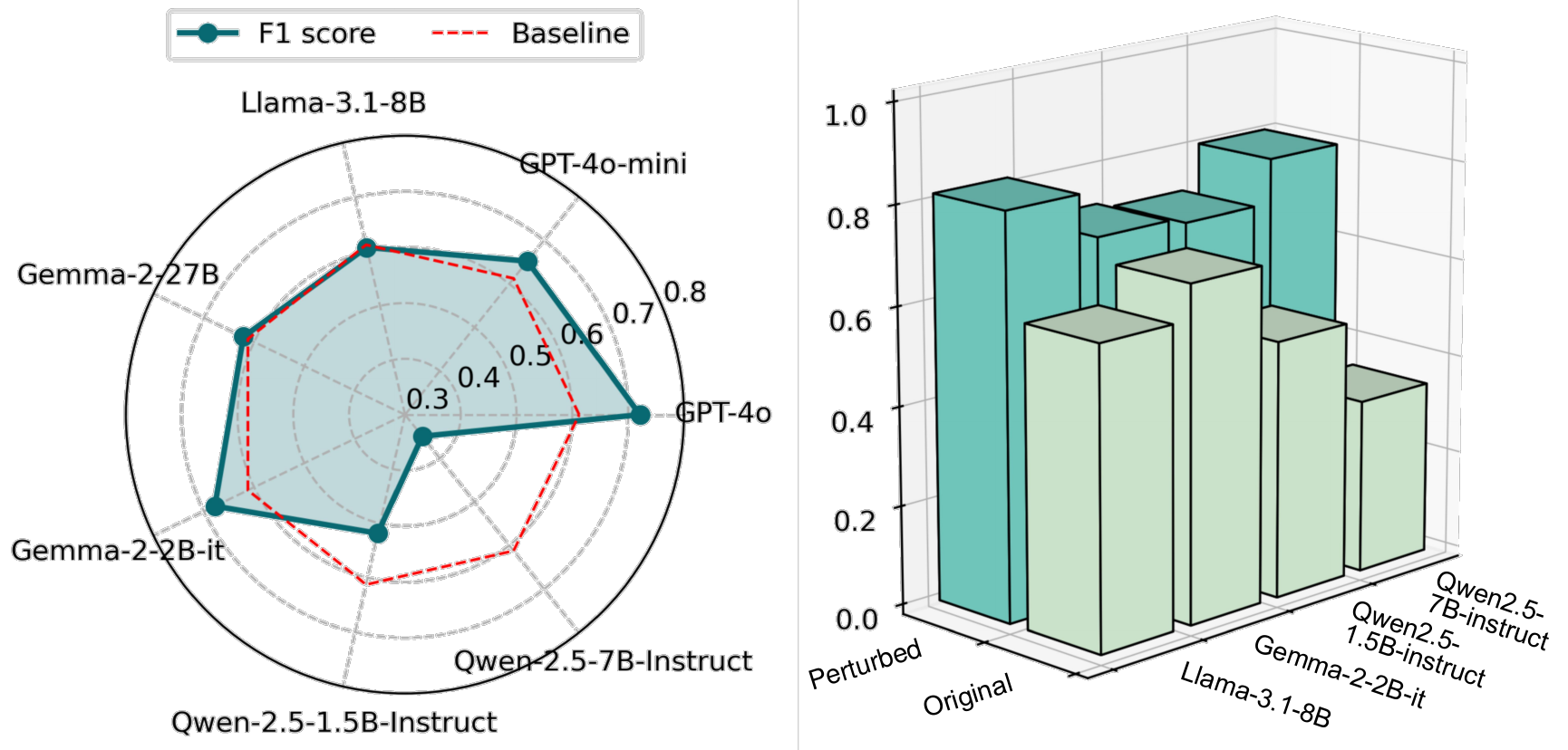}
        \vspace{-12pt}
        \captionof{figure}{Comparison of classification performance using $F_{0.5}$ Scores. \textbf{Left:} $F_{0.5}$ scores of seven prompt-based classifiers compared to the baseline without a classifier (recall is 1 when all problems are enhanced directly). \textbf{Right:} $F_{0.5}$ scores of four fine-tuned classifiers after training, showing significant improvements over prompt-based classifiers.}
        \label{fig:f1_score}
    \end{minipage}
    \hfill
    \begin{minipage}[b]{0.48\linewidth}
        \centering
        \includegraphics[width=\linewidth]{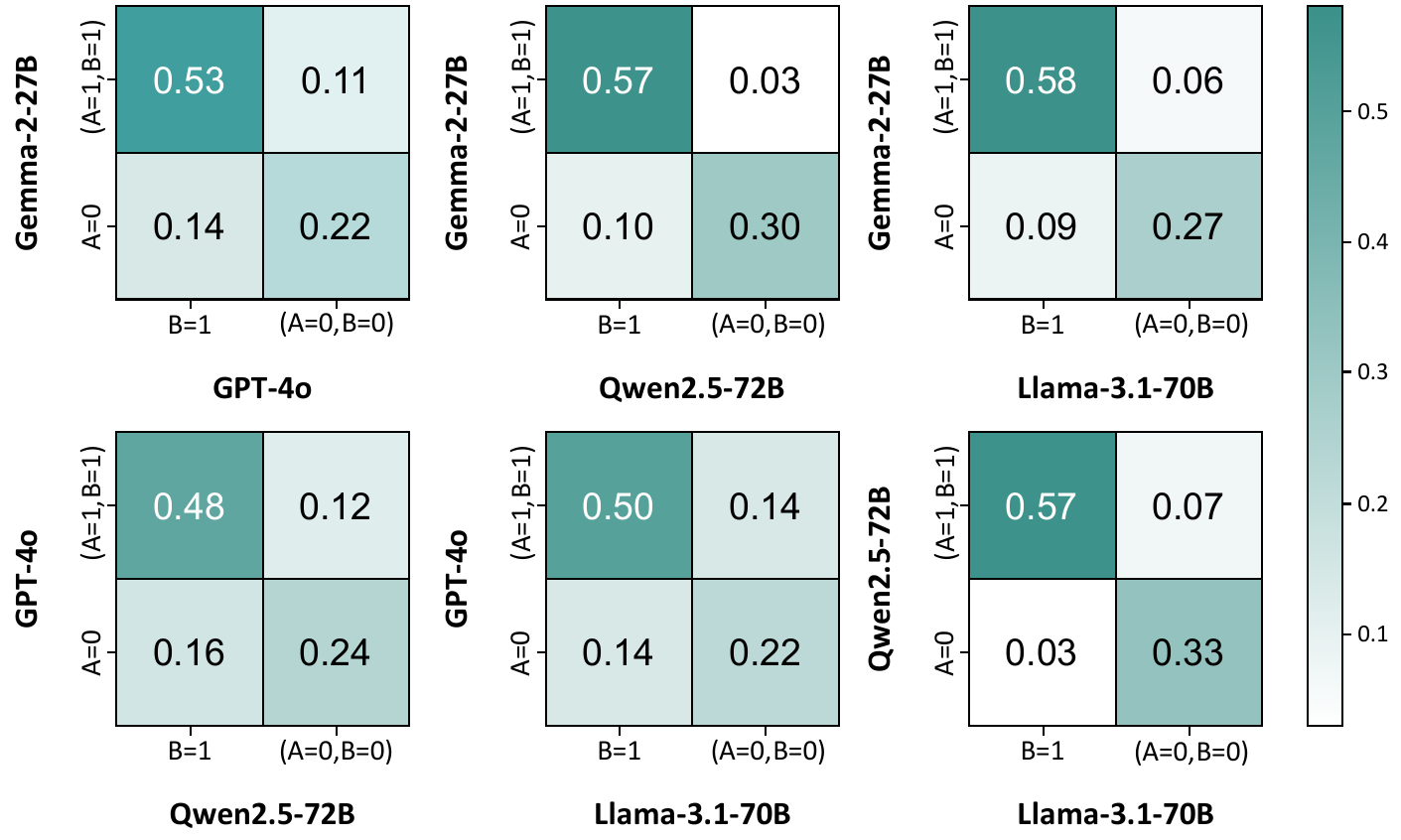}
        \vspace{-12pt}
        \captionof{figure}{The result of whether the samples are perturbable by two models, A and B. Here, A=1 indicates that the sample is easy-to-perturb for model A, while A=0 means it is hard-to-perturb for model A. The numbers in each cell represent the percentage of samples in each category.}
        \label{fig:classifier_matrix}
    \end{minipage}
\end{figure}

To evaluate the generalizability of the classifier, we examine whether perturbation difficulty is consistent across different LLMs. As shown in the confusion matrix in \autoref{fig:classifier_matrix}, there is strong alignment between models: around 82\% of questions are either distractable or non-distractable for both models in any pairwise comparison. This consistency suggests that model-agnostic classifiers can be trained to identify distractable inputs.

We implement two types of classifiers: (1) \textit{Prompt-based classifiers}, which leverage LLM Judge \citep{ye2024justice, wang2025trusteval}, and (2) \textit{Fine-tuned classifiers}, trained on 1,080 annotated examples with 120 held-out test cases. As shown in \autoref{tab:cross_model_precision}, classifiers trained on data from GPT-4o-mini generalize effectively to stronger models, maintaining high precision across model families. We use the F$_\beta$ score with $\beta=0.5$ to prioritize precision. The formal definition is provided in Appendix~\ref{appendix:experiment_settings}.

As illustrated in \autoref{fig:f1_score} and \autoref{tab:classifier_results}, fine-tuned classifiers significantly improve overall efficiency by accurately filtering out hard-to-perturb samples. In particular, our best fine-tuned classifiers achieve up to 83\% precision on identifying distractable problems, outperforming the best prompt-based baseline (GPT-4o) at 68\%. The reduction in wasted computational effort and improvement in overall perturbation success rates will be analyzed in more detail in Experiment~\ref{sec:ablation_study}.

To further assess whether the classifier generalizes across tasks, we conduct a cross-dataset evaluation.
Specifically, we train the classifier using three datasets (MMLU, CommonsenseQA,
and OpenbookQA), and validate its performance on the held-out TruthfulQA.
Table~\ref{tab:clf_cross_dataset} presents the $F_{0.5}$ scores under this setting.
Despite not being trained on TruthfulQA, the classifier maintains comparable performance,
suggesting that it captures general signals of distraction susceptibility rather than dataset-specific patterns.

\begin{wraptable}{r}{0.52\linewidth}
\vspace{-1em}
\centering
\scriptsize

\definecolor{rowlight}{HTML}{F8FAFB}
\definecolor{rowdark}{HTML}{EEF3F5}
\definecolor{headerbg}{HTML}{DCE7EF}
\definecolor{upclr}{HTML}{267300}
\definecolor{downclr}{HTML}{B6462C}
\caption{The impact of the classifier on the perturbation success rate of the LLMs. The full model names are: GPT-4o, Gemma-2-27B, Llama-3.1-70B, and Qwen2.5-72B. The rows display the perturbation success rate with and without the classifier.}
\label{tab:cross_model_precision}

\rowcolors{2}{rowlight}{rowdark}
\begin{tabular}{
    >{\columncolor{headerbg}}l
    >{\columncolor{headerbg}}c
    >{\columncolor{headerbg}}c
    >{\columncolor{headerbg}}c
    >{\columncolor{headerbg}}c
}
    \toprule
    \textbf{Mode} & \textbf{GPT-4o} & \textbf{Gemma-2} & \textbf{Llama-3.1} & \textbf{Qwen2.5} \\
    \midrule
    w/o classifier & 0.527 & 0.592 & 0.581 & 0.563 \\
    w/ classifier  & 0.723 & 0.791 & 0.754 & 0.735 \\
    \bottomrule
\end{tabular}
\vspace{-1em}
\end{wraptable}

% The optimized pipeline reduces the computational resources required by 37\% through the reduction of the targeted search space, demonstrating that the selection of problem guided by the classifier maintains adversarial effectiveness while significantly improving scalability.

\subsection{Ablation Study}
\label{sec:ablation_study}

\textbf{Impact of the value function.} We evaluate the effectiveness of the designed value function \(\mathcal{V}(P')\) and its components introduced in Method~\ref{sec:value_function} for guiding the search process. This function incorporates two key factors: the success rate \( r_M(P') \) of the victim model on the distracted version of the problem \( P' \), and a depth penalty \( \text{depth}^{-\gamma} \), which helps balance the trade-off between exploration and computational efficiency.

To assess the individual contributions of these components, we randomly selected questions from four datasets and measured the distraction success rate (defined as the percentage of questions for which the model fails completely under distraction, i.e., \( r_M(P') = 0 \)). The results show that using the complete value function yields a 59\% success rate. Removing the depth penalty reduces this rate to 57\%, and removing the success rate term further decreases it to 53\%. These findings confirm that both the model failure signal and the search depth control play important roles in identifying effective adaptive distractions.

\textbf{Cost saved by classifier.} To examine the impact of the classifier, we compared performance with and without classifier filtering. In this experiment, we randomly selected questions from four datasets. In the classifier condition, we used our classifier to select 100 questions predicted to be susceptible to distraction. In the baseline condition, we sampled 100 questions at random without any filtering.

\begin{table*}[t]
\centering
\definecolor{rowlight}{HTML}{F8FAFB}
\definecolor{rowdark}{HTML}{EEF3F5}
\definecolor{headerbg}{HTML}{DCE7EF}
\definecolor{upclr}{HTML}{267300}
\definecolor{downclr}{HTML}{B6462C}
\newcommand{\up}[1]{\textcolor{upclr}{\tiny$\uparrow$#1}}
\newcommand{\down}[1]{\textcolor{downclr}{\tiny$\downarrow$#1}}

\renewcommand{\arraystretch}{1.2}
\caption{Computational cost comparison with and without classifier. \textbf{Inp. Tok.}: Number of input tokens, \textbf{Out. Tok.}: Number of output tokens, \textbf{Pert. Ques.}: Number of successfully distracted questions, \textbf{Pert. Rate (\%)}: Distraction success rate, \textbf{Cost/Ques. (\$)}: USD cost per successfully distracted question.}
\label{tab:classifier_comparison}

\rowcolors{2}{rowlight}{rowdark}
\begin{tabular}{
    >{\columncolor{headerbg}}l
    >{\columncolor{headerbg}}c
    >{\columncolor{headerbg}}c
    >{\columncolor{headerbg}}c
    >{\columncolor{headerbg}}c
    >{\columncolor{headerbg}}c
}
    \toprule
    \textbf{Mode} & \textbf{Inp. Tok.} & \textbf{Out. Tok.} & \textbf{Pert. Ques.} & \textbf{Pert. Rate (\%)} & \textbf{Cost (\$)} \\
    \midrule
    \textit{w/o} classifier & 3.69M & 1.48M & 175 & 59\% & 0.0082 \\
    \textit{w/} classifier  & 3.81M & 1.57M & 236 & 82\%\up{38.9\%} & 0.0064\down{21.9\%} \\
    \bottomrule
\end{tabular}
\vspace{-1em}
\end{table*}

As shown in \autoref{tab:classifier_comparison}, incorporating the classifier significantly improves both the effectiveness and the efficiency of the generation process. By filtering out hard-to-distract questions, the classifier enables better allocation of resources to more promising inputs. This leads to a 38.9\% increase in success rate and a 21.9\% reduction in average cost per successful distraction.

\subsection{Mitigation under Adaptive Distraction}

Notably, models that perform well on original questions but fail on distracted ones demonstrate that they possess the necessary knowledge yet remain vulnerable to contextual interference. To address this issue, we explored both \textbf{prompt-based (i.e., training-free)} and \textbf{training-based strategies} to improve their performance on these challenging questions.

\begin{wraptable}{r}{0.5\linewidth}
\vspace{-1em}
\centering
\scriptsize
\setlength{\tabcolsep}{4mm}

\definecolor{rowlight}{HTML}{F8FAFB}
\definecolor{rowdark}{HTML}{EEF3F5}
\definecolor{headerbg}{HTML}{DCE7EF}

\caption{Model accuracy before and after prompt-based mitigation (\textbf{Orig.} vs \textbf{Enh.}).}
\label{tab:prompt_enhanced}

\rowcolors{2}{rowlight}{rowdark}
\begin{tabular}{
    >{\columncolor{headerbg}}l
    >{\columncolor{headerbg}}r
    >{\columncolor{headerbg}}r
    >{\columncolor{headerbg}}r
}
    \toprule
    \textbf{Model} & \textbf{Orig.} & \textbf{Enh.} & \textbf{Diff.} \\
    \midrule
    GPT-4o-mini       & 0.185 & 0.211 & \textcolor{green!40!black}{$+$0.026} \\
    Llama-3.1-8B      & 0.247 & 0.251 & \textcolor{green!40!black}{$+$0.003} \\
    Gemma-2-27B       & 0.237 & 0.255 & \textcolor{green!40!black}{$+$0.018} \\
    o1-mini           & 0.374 & 0.366 & \textcolor{red!40!black}{$-$0.008} \\
    Qwen2.5-72B       & 0.334 & 0.343 & \textcolor{green!40!black}{$+$0.009} \\
    GPT-4o            & 0.390 & 0.391 & \textcolor{green!40!black}{$+$0.002} \\
    Claude-3.5-sonnet & 0.495 & 0.481 & \textcolor{red!40!black}{$-$0.013} \\
    \bottomrule
\end{tabular}
\vspace{-1em}
\end{wraptable}

\textbf{Prompt-based mitigation is unable to alleviate distraction vulnerability.} Since our adaptive distractions preserve the core question semantics while introducing task-irrelevant context, we tested whether adding explicit instructions in the prompt could help models focus on the essential content and ignore misleading information. Specifically, we modified the original prompts to include guidance on identifying and prioritizing key components of the question. Detailed templates are provided in Appendix~\ref{appendix:prompt_template}. As shown in \autoref{tab:prompt_enhanced}, this approach yielded only marginal improvements. Some models, such as o1-mini and Claude-3.5-Sonnet, even showed slightly lower accuracy after prompt modifications. This suggests that the contextual interference introduced by adaptive distractions cannot be effectively mitigated through prompt refinement alone.

We also tested additional prompting methods that have shown promise in prior work, including In-context learning~\cite{wei2022chain} and Self-consistency~\cite{wang2022self}. As shown in Appendix~\ref{app:prompt_variants},
none of these approaches substantially recover the original performance when faced with adaptive distractions. This reinforces our finding that prompt-based methods alone offer limited robustness.

\begin{wraptable}{r}{0.5\linewidth}
\vspace{-1em}
\centering
\scriptsize
\setlength{\tabcolsep}{3mm}

\definecolor{rowlight}{HTML}{F8FAFB}
\definecolor{rowdark}{HTML}{EEF3F5}
\definecolor{headerbg}{HTML}{DCE7EF}

\caption{Model accuracy before and after DPO training. \textbf{Retain} shows the fraction of original incorrect answers that remain incorrect after training.}
\label{tab:dpo_enhanced}

\rowcolors{2}{rowlight}{rowdark}
\begin{tabular}{
    >{\columncolor{headerbg}}l
    >{\columncolor{headerbg}}c
    >{\columncolor{headerbg}}c
    >{\columncolor{headerbg}}c
    >{\columncolor{headerbg}}c
}
    \toprule
    \textbf{Model} & \textbf{Orig.} & \textbf{Enh.} & \textbf{Diff.} & \textbf{Retain} \\
    \midrule
    Gemma-2-2B       & 0.257 & 0.432 & \textcolor{green!40!black}{+0.175} & 0.788 \\
    Qwen2.5-7B       & 0.212 & 0.440 & \textcolor{green!40!black}{+0.228} & 0.763 \\
    Phi-3.5-mini     & 0.195 & 0.680 & \textcolor{green!40!black}{+0.485} & 0.821 \\
    \midrule
    \textit{GPT-4o}           & 0.568 & --     & --                                  & --    \\
    \textit{Qwen2.5-72B}      & 0.519 & --     & --                                  & --    \\
    \textit{GPT-4o-mini}      & 0.232 & --     & --                                  & --    \\
    \bottomrule
\end{tabular}
\vspace{-1em}
\end{wraptable}

As shown in \autoref{tab:dpo_enhanced}, all three models benefited significantly from training. The Phi-3.5-mini model, in particular, achieved a post-training accuracy that surpassed even GPT-4o on the same distracted inputs. Detailed case studies in Appendix~\ref{appendix:case_study} show that the improvements were not merely due to new knowledge acquisition. Rather, fine-tuned models showed better focus on relevant question content and stronger resistance to irrelevant distractions. A large fraction of the originally incorrect answers remained incorrect, such as 82.1\% for Phi-3.5-mini, suggesting that performance gains came from improved robustness rather than memorization. Additional analysis in Appendix~\ref{app:sft_dpo} confirms that DPO offers greater gains than supervised fine-tuning, while preserving performance on clean questions, which further validates that our improvements stem from true robustness rather than memorization.

%% file: Section/5-conclusion.tex
\section{Conclusion}
In this work, we propose a framework to assess the contextual robustness of language models by generating adaptive distractions, which are semantically coherent but task-irrelevant additions. Our tree-based search method produces challenging examples that induce consistent performance drops across models and datasets. Among mitigation strategies, post-training methods such as DPO offer the most reliable improvements. Ultimately, our approach offers a scalable tool for evaluating and improving LLM reliability in real-world applications. Future work will integrate our distraction generation into training loops to further strengthen contextual robustness.

% \clearpage

%% file: Section/99-appendix.tex
\appendix

\section{Related Work}

\subsection{Contextual Robustness in LLMs}

Recent studies have highlighted that LLMs often fail when presented with semantically coherent yet task-irrelevant contextual information. Shi et al.~\cite{shi2023large} introduced a benchmark demonstrating that irrelevant context can severely degrade LLM accuracy in arithmetic reasoning tasks. Liu et al.~\cite{liu2023lost} showed that model performance significantly drops when key information is placed in the middle of long contexts, indicating positional sensitivity in attention mechanisms. Similar findings have emerged in narrative distraction scenarios~\cite{chatziveroglou2025exploring}, misprimed probe studies~\cite{kassner2019negated}, and irrelevant document retrieval contexts~\cite{yoran2023making}. Collectively, these works underscore contextual distraction as a prevalent vulnerability affecting modern LLMs.

\subsection{Generation of Contextual Perturbations}

To systematically probe LLM sensitivity to irrelevant context, various methods have been developed to generate targeted perturbations. Zhu et al.~\cite{zhu2024dynamic} proposed a dynamic evaluation approach using meta probing agents that restructure tasks to surface latent weaknesses in model behavior. Similarly, optimization-based prompt injection methods have been explored to create adversarial inputs aimed at exploiting model biases or alignment issues~\cite{shi2024optimization, paulus2024advprompter}. Chatziveroglou et al.~\cite{chatziveroglou2025exploring} further validate that even semantically coherent but irrelevant narratives can significantly reduce LLM accuracy. Despite these efforts, existing perturbation generation techniques typically focus on altering prompts or instructions without necessarily preserving answer correctness or targeting semantic coherence explicitly.
% In contrast, our approach generates adaptive, semantically coherent contextual perturbations through tree search, specifically designed to reveal LLM vulnerabilities while preserving the original question and answer structure.

\subsection{Mitigation Strategies}

Several approaches have attempted to address the issue of contextual distraction by enhancing model robustness \cite{jiang2024enhancing}. PromptBreeder~\cite{fernando2023promptbreeder} employs evolutionary strategies to optimize task prompts, implicitly strengthening model robustness against perturbations. Wu et al.~\cite{wu2024instructing} introduced prompting strategies instructing models to explicitly ignore irrelevant information, and Wang et al.~\cite{wang2022self} showed that self-consistency decoding can improve reliability by aggregating predictions from multiple reasoning paths. Zhu et al.~\cite{zhu2025focus} explored attention mechanisms, identifying internal attention directions to guide models toward more relevant context. Moreover, Yoran et al.~\cite{yoran2023making} demonstrated the effectiveness of fine-tuning on mixed-relevance data to improve robustness in retrieval-augmented scenarios. Direct preference optimization (DPO)~\cite{rafailov2023direct}, a targeted fine-tuning technique, has gained particular attention for its effectiveness in enhancing model alignment and resilience to distracting inputs. While prompt-based approaches remain attractive due to their simplicity, our findings indicate they have limited efficacy against adaptive distractions. Our experiments highlight the stronger robustness achieved by fine-tuning strategies, validating the effectiveness of targeted mitigation approaches in addressing contextual distraction.

\section{Experiment Details}
\label{appendix:settings}

% \subsection{Models}
% \label{appendix:model_settings}
% As shown in \autoref{tab:all_models}, we used four high-performance proprietary models in our experiments: GPT-4o \citep{hurst2024gpt}, GPT-4o-mini \citep{openai2024gpt4omini}, Claude-3.5-Sonnet \citep{anthropic2024claude35}, and o1-mini \citep{jaech2024openai}. In addition, we included eight open-weight models: Gemma-2-2B, Gemma-2-27B \citep{gemma_2024}, Qwen2.5-1.5B, Qwen2.5-7B, Qwen2.5-72B \citep{qwen2,qwen2.5}, Llama-3.1-8B \citep{meta2024llama31_8b}, Llama-3.1-70B \citep{meta2024llama31_70b} and Phi-3.5-mini \citep{abdin2024phi}.

\definecolor{headerbg}{RGB}{44,62,80}
\definecolor{rowgray}{RGB}{245,245,245}
\definecolor{rowblue}{RGB}{234,242,248}

\begin{table}[htbp]
\vspace{-1em}
\centering
\renewcommand{\arraystretch}{1}
\setlength{\tabcolsep}{1.4mm}
\caption{Models used in our experiments along with their versions, organizations, licenses, and purposes. \textit{Gen}: Model used for generating questions (as a proxy or victim); \textit{Eval}: Model used for evaluating datasets; \textit{Clf}: Model used as a classifier to filter questions.}
\label{tab:all_models}
\rowcolors{2}{rowgray}{white}  % Alternate row colors
\scalebox{0.8}{
\begin{tabular}{lcccccc}
    \toprule[1.5pt]
    \rowcolor{headerbg}
    \textcolor{white}{\textbf{Model}} & 
    \textcolor{white}{\textbf{Version}} & 
    \textcolor{white}{\textbf{Organization}} & 
    \textcolor{white}{\textbf{License}} & 
    \textcolor{white}{\textbf{Gen}} & 
    \textcolor{white}{\textbf{Eval}} & 
    \textcolor{white}{\textbf{Clf}} \\
    \midrule[0.8pt]
    GPT-4o-mini       & gpt-4o-mini-2024-07-18       & OpenAI      & Proprietary            & \textcolor{red}{\checkmark} & \textcolor{red}{\checkmark} & \\
    GPT-4o            & gpt-4o-2024-08-06            & OpenAI      & Proprietary            & \textcolor{red}{\checkmark} & \textcolor{red}{\checkmark} & \\
    Gemma-2-2B        & Gemma-2-2B-it                & Google      & Gemma License          &  & \textcolor{red}{\checkmark} & \textcolor{red}{\checkmark} \\
    Gemma-2-27B       & Gemma-2-27B-it               & Google      & Gemma License          & \textcolor{red}{\checkmark} & \textcolor{red}{\checkmark} & \\
    Llama-3.1-8B      & Meta-Llama-3.1-8B-Instruct   & Meta        & Llama 3.1 Community    &  & \textcolor{red}{\checkmark} & \textcolor{red}{\checkmark} \\
    Llama-3.1-70B     & Meta-Llama-3.1-70B-Instruct  & Meta        & Llama 3.1 Community    & \textcolor{red}{\checkmark} & \textcolor{red}{\checkmark} & \\
    Qwen2.5-1.5B      & Qwen2.5-1.5B-Instruct        & Alibaba     & Qwen License           &  &  & \textcolor{red}{\checkmark} \\
    Qwen2.5-7B        & Qwen2.5-7B-Instruct          & Alibaba     & Qwen License           &  & \textcolor{red}{\checkmark} & \textcolor{red}{\checkmark} \\
    Qwen2.5-72B       & Qwen2.5-72B-Instruct         & Alibaba     & Qwen License           & \textcolor{red}{\checkmark} & \textcolor{red}{\checkmark} & \\
    o1-mini           & o1-mini-2024-09-12           & OpenAI      & Proprietary            &  & \textcolor{red}{\checkmark} & \\
    Phi-3.5-mini      & Phi-3.5-mini-instruct        & Microsoft   & MIT                    &  & \textcolor{red}{\checkmark} & \textcolor{red}{\checkmark} \\
    Claude-3.5-Sonnet & claude-3-5-sonnet-20241022 & Anthropic   & Proprietary            &  & \textcolor{red}{\checkmark} & \\
    \bottomrule[1.5pt]
\end{tabular}}
\vspace{-10pt}
\end{table}

\subsection{Experiment Settings}
\label{appendix:experiment_settings}
In all experiments, we adopt the same parameter settings. Specifically, we set the length threshold \(\lambda = 10\), the semantic threshold \(\tau_C = 0.5\), the number of simulation times \(n = 5\), and the diversity limit \(n_1 = 3\). Additionally, we use the same model as both the proxy model and the victim model.

\textbf{Experimental details of different victim models.} We selected five victim models with varying capabilities: GPT-4o, GPT-4o-mini, Llama-3.1-70B, Qwen2.5-72B, and Gemma-2-27B. From each of the four datasets, namely MMLU, CommonsenseQA, OpenbookQA, and TruthfulQA, we randomly sampled 100 original questions. Each victim model enhanced these questions via our search framework, creating five distinct enhanced datasets. To evaluate the effectiveness of these enhanced questions, we tested the performance of seven different models: GPT-4o-mini, Gemma-2-27B, Llama-3.1-8B, Qwen2.5-72B, o1-mini, GPT-4o, and Claude-3.5-Sonnet. All models were evaluated using a zero-shot approach with CoT prompting templates. This setup allowed us to systematically analyze the relationship between victim model capability and the difficulty of the generated enhanced questions.
The results of this experiment are summarized in \autoref{fig:multi_results}.

\begin{figure*}[htbp]
    \centering
    \includegraphics[width=\linewidth]{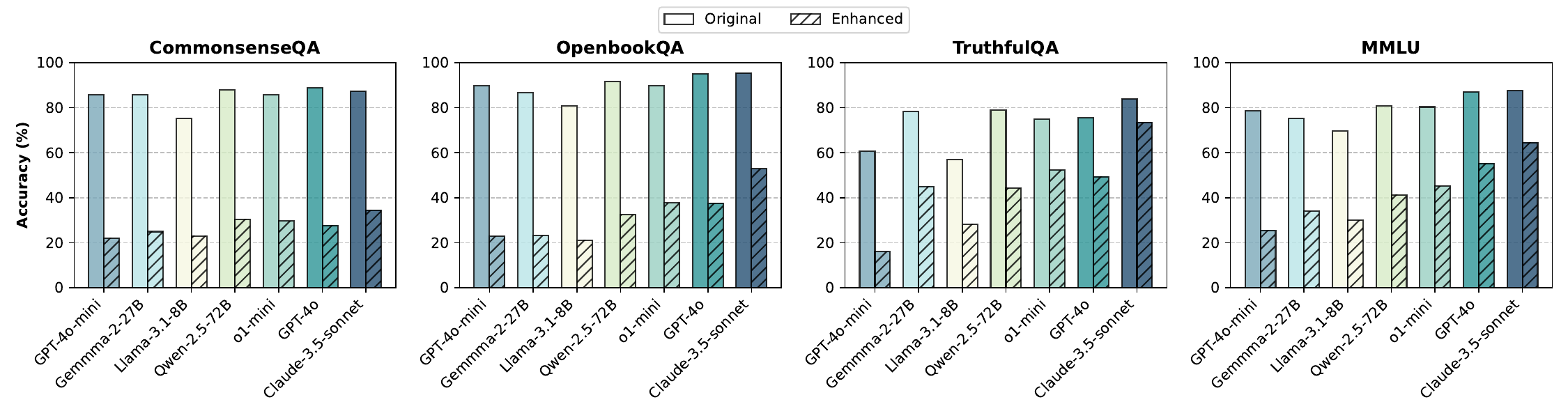}
    \vspace{-2em}
    \caption{Overall results between 4 datasets.}
    \label{fig:main_results}
\end{figure*}

\textbf{Experimental details of scale-up experiment.} We selected GPT-4o-mini as the victim model for question enhancement due to its balance between perturbation effectiveness and computational efficiency. From the same four datasets, we sampled 300 questions per dataset, resulting in a total of 1200 original questions. Similar to the first experiment, the enhanced questions were tested across the same seven models: GPT-4o-mini, Gemma-2-27B, Llama-3.1-8B, Qwen2.5-72B, o1-mini, GPT-4o, and Claude-3.5-Sonnet. All evaluations were conducted using zero-shot CoT prompting templates. This larger-scale experiment provided a more comprehensive analysis of the generalizability of our perturbation methodology.
The results of this experiment are summarized in \autoref{fig:main_results} and \autoref{tab:main_results}.

\textbf{Experimental details of baseline methods.} To validate the effectiveness of our tree-based search framework, we implemented two baseline perturbation approaches for comparison. The \textbf{Irrelevant Context Augmentation (ICA)} method performed semantic-preserving length augmentation by expanding original questions with task-irrelevant but semantically coherent contextual information, such as explanatory clauses or redundant details. The \textbf{Single-Prompt Distraction (SPD)} baseline utilized our perturbation prompt template (details in \autoref{appendix:prompt_template}) through Claude-3.5-Sonnet for automatic distraction generation without subsequent search optimization. For a fair comparison, all baseline methods processed the same 100 original questions from four datasets using Claude-3.5-Sonnet as the executor. The enhanced questions were evaluated under identical zero-shot CoT settings across seven target models. This demonstrates the crucial role of our tree-based search mechanism in identifying optimal perturbation combinations rather than relying on simple length expansion or single-pass prompt perturbations.
The results of this experiment are summarized in \autoref{tab:baseline}.

\textbf{Experimental details of classifier.} We used  1200 original questions from 4 datasets, splitting them into training, test, and validation sets. Specifically, 80 percent of the data was allocated to training, with 10 percent of the training set reserved for validation, and the remaining 10 percent was used for testing. For the prompt-based classifiers, we designed specific prompts to guide the models in determining whether a problem was hard to perturb. We evaluated the classification performance of seven models: GPT-4o-mini, GPT-4o, Llama-3.1-8B, Gemma-2-27B, Gemma-2-2B, Qwen2.5-1.5B, and Qwen2.5-7B. A baseline configuration without any classifier was also included for comparison. The effectiveness of these classifiers was measured using the F1-score with beta equal to 0.5, which prioritizes precision over recall. For the training-based classifiers, we used supervised fine-tuning with LoRA on four open-source models: Llama-3.1-8B, Gemma-2-2B, Qwen2.5-1.5B, and Qwen2.5-7B. The training was conducted on a single RTX 4090 GPU, with a learning rate set to 1e-4 and a total of five epochs. The performance of these fine-tuned classifiers was also evaluated using the F1-score on the test set. This experimental design allowed us to compare the utility of prompt-based and training-based classifiers in identifying hard-to-perturb questions. The results of this experiment are summarized in \autoref{fig:classifier_matrix}, \autoref{tab:cross_model_precision}, \autoref{fig:f1_score} and \autoref{tab:classifier_results}.

To evaluate the tradeoff between precision and recall in our classifier analysis, we report the F$_\beta$ score with $\beta = 0.5$. This metric places greater emphasis on precision, which is desirable in our use case. The score is defined as:

\begin{equation}
\text{F}_{\beta} = (1 + \beta^2) \times \frac{\text{Precision} \times \text{Recall}}{(\beta^2 \times \text{Precision}) + \text{Recall}}.
\end{equation}

\begin{table}[htbp]
\centering
\small
\setlength{\tabcolsep}{2.5mm}

\definecolor{rowlight}{HTML}{F8FAFB}
\definecolor{rowdark}{HTML}{EEF3F5}

\caption{Performance of the classifier under Prompt-Based and Fine-Tuned methods. The table reports Precision, Recall, and F\textsubscript{0.5} scores for both Prompt-Based (left) and Fine-Tuned (right) classifiers. Fine-Tuned models are marked in the Fine-Tuned columns. Baseline represents the performance of the system without using classifier.}
\label{tab:classifier_results}

\rowcolors{3}{rowlight}{rowdark}
\begin{tabular}{l|ccc|ccc}
    \toprule
    \multirow{2}{*}{\textbf{Model}} & \multicolumn{3}{c|}{\textbf{Prompt-Based}} & \multicolumn{3}{c}{\textbf{Fine-Tuned}} \\
     & \textbf{Precision} & \textbf{Recall} & \textbf{F\textsubscript{0.5}} & \textbf{Precision} & \textbf{Recall} & \textbf{F\textsubscript{0.5}} \\
    \midrule
    GPT-4o-mini      & 0.606 & 0.940 & 0.652 & --    & --    & --    \\
    GPT-4o           & \textbf{0.685} & 0.910 & \textbf{0.721} & --    & --    & --    \\
    Llama-3.1-8B     & 0.555 & 0.985 & 0.608 & \textbf{0.812} & 0.836 & \textbf{0.816} \\
    Gemma-2-27B      & 0.568 & 1.000 & 0.622 & --    & --    & --    \\
    Gemma-2-2B       & 0.558 & 0.866 & 0.678 & 0.712 & 0.776 & 0.724 \\
    Qwen2.5-1.5B     & 0.534 & 0.463 & 0.518 & 0.719 & 0.687 & 0.712 \\
    Qwen2.5-7B       & 0.526 & 0.149 & 0.350 & 0.797 & 0.821 & 0.802 \\
    \midrule
    Baseline         & 0.558 & 1.000 & 0.612 & 0.558 & 1.000 & 0.612 \\
    \bottomrule
\end{tabular}
\end{table}

\textbf{Experimental details of mitigation.} We curated approximately 1200 preference data pairs. Each preference pair consisted of a question, a correct answer, and an incorrect answer collected from model responses in prior experiments. To ensure a fair evaluation, we guaranteed that enhanced questions originating from the same original question did not appear in both the training and test sets. The data was split into training, validation, and test sets, with 80 percent of the data used for training, 10 percent of the training set reserved for validation, and 20 percent allocated to testing. For prompt-based enhancement, we designed new prompt templates aimed at improving model focus on the core question content and tested them on seven models: GPT-4o-mini, Gemma-2-27B, Llama-3.1-8B, Qwen2.5-72B, o1-mini, GPT-4o, and Claude-3.5-Sonnet. For training-based enhancement, we fine-tuned three open-source models, namely Gemma-2-2B, Qwen2.5-7B, and Phi-3.5-mini. Using the Direct Preference Optimization algorithm, the fine-tuning was performed on two RTX 4090 GPUs with a learning rate set to 2e-4 and five epochs. The preference loss was implemented with a sigmoid activation function. The fine-tuned models were evaluated against three high-performance baseline models, specifically GPT-4o, GPT-4o-mini, and Qwen2.5-72B, using the original zero-shot with CoT prompting templates on the test set. This experiment provided insights into the effectiveness of both prompt-based and training-based approaches in improving model robustness against enhanced questions. The results of this experiment are summarized in \autoref{tab:prompt_enhanced} and \autoref{tab:dpo_enhanced}.

\subsection{Preliminary Experimental Results}
\label{app:preliminary_results}

To support our claim in the introduction regarding the limited effectiveness of static distraction methods, we present a comparison of performance drops induced by our adaptive distraction framework  and a representative static method, GSM-IC \cite{shi2023large}, in \autoref{tab:preliminary_results}. The results demonstrate that while GSM-IC causes minimal performance drops (average 1.8\%) on advanced models, our method achieves significantly larger drops (average 45\%), highlighting its potency in challenging LLM contextual robustness.

\begin{table}[htbp]
\centering

\definecolor{rowlight}{HTML}{F8FAFB}
\definecolor{rowdark}{HTML}{EEF3F5}
\definecolor{headerbg}{HTML}{DCE7EF}

\caption{Comparison of performance drops (\%) on GSM-IC \cite{shi2023large} and our adaptive distraction.}
\label{tab:preliminary_results}

\rowcolors{2}{rowlight}{rowdark}
\resizebox{1\textwidth}{!}{%
\begin{tabular}{
    >{\columncolor{headerbg}}l
    >{\columncolor{headerbg}}c
    >{\columncolor{headerbg}}c
    >{\columncolor{headerbg}}c
    >{\columncolor{headerbg}}c
    >{\columncolor{headerbg}}c
    >{\columncolor{headerbg}}c
}
\toprule
\textbf{Model} & \textbf{GPT-4o-mini} & \textbf{GPT-4o} & \textbf{Qwen2.5-72B} & \textbf{Gemma-2-27B} & \textbf{Claude-3.5-sonnet} & \textbf{o1-mini} \\
\midrule
GSM-IC \cite{shi2023large} & 4.1 & 1.2 & 1.4 & 2.2 & 0.8 & 1.2 \\
\bottomrule
\end{tabular}%
}
\vspace{-5pt}
\end{table}

\subsection{Experiment Analysis}
\begin{figure}[htbp]
    \centering
    \includegraphics[width=0.5\linewidth]{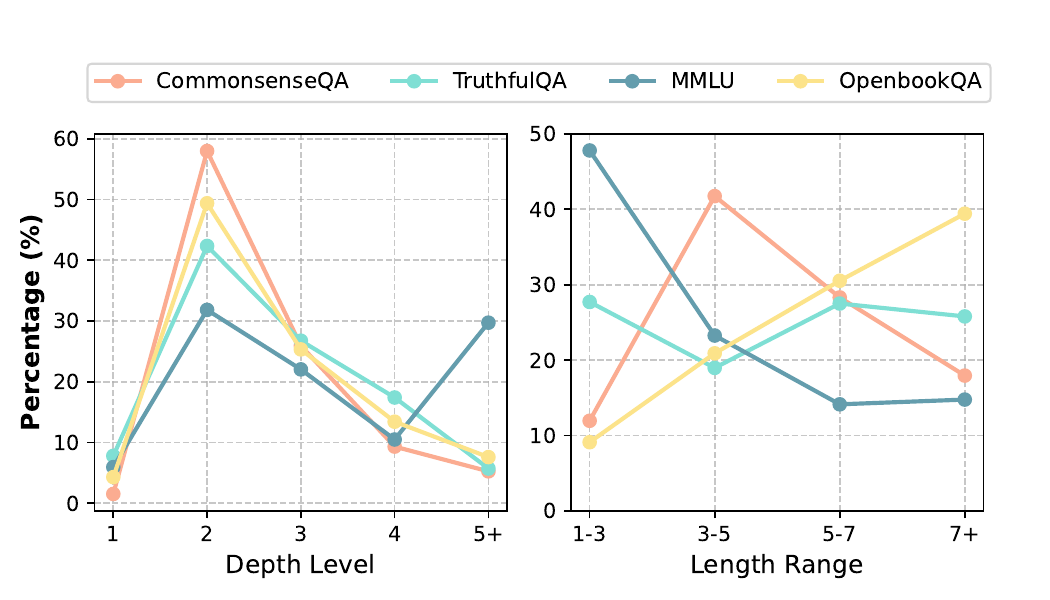}
    \vspace{-10pt}
    \caption{Distribution analysis of perturbation chain depth and enhanced question length ratio across four datasets.}
    \label{fig:length_depth}
\end{figure}

\textbf{Distribution Analysis of Enhanced Questions.} Our analysis of the search process reveals interesting patterns in both the depth of perturbation chains and the length ratios of enhanced questions across different datasets. As shown in \autoref{fig:length_depth}, the majority of successful perturbations were found at relatively shallow depths, particularly for CommonsenseQA and OpenbookQA, where approximately 85\% and 80\% of effective perturbations were discovered within the first three levels. However, MMLU exhibited a notably different pattern, with nearly 30\% of perturbations requiring five or more steps to achieve effectiveness. This suggests that questions testing specialized knowledge often require more sophisticated and layered perturbations to successfully challenge model performance. The length ratios of enhanced questions also varied significantly across datasets. OpenbookQA showed a tendency toward longer perturbations, with about 70\% of enhanced questions being more than five times longer than their original versions. In contrast, MMLU questions maintained relatively compact perturbations, with nearly half of the enhanced questions staying within three times the original length. These distributions reflect the varying complexity required to effectively perturb different types of questions and highlight how the nature of the underlying task influences the perturbation process.

\subsection{Cross-Dataset Generalization of the Classifier}

We present additional evaluation of the classifier’s ability to generalize across datasets.
As shown in Table~\ref{tab:clf_cross_dataset}, the $F_{0.5}$ score remains stable even
when the classifier is tested on a domain it was not trained on.

\begin{table}[h]
\centering
\small

\definecolor{rowlight}{HTML}{F8FAFB}
\definecolor{rowdark}{HTML}{EEF3F5}
\definecolor{headerbg}{HTML}{DCE7EF}

\caption{$F_{0.5}$ scores of the classifier trained on MMLU, CommonsenseQA, and OpenbookQA, and tested on the held-out TruthfulQA dataset.}
\label{tab:clf_cross_dataset}

\rowcolors{2}{rowlight}{rowdark}
\begin{tabular}{
    >{\columncolor{headerbg}}l
    >{\columncolor{headerbg}}c
    >{\columncolor{headerbg}}c
}
\toprule
\textbf{Model} & \textbf{Orig. Test} & \textbf{Cross-Dataset} \\
\midrule
Gemma-2-2B     & 0.724 & 0.735 (+0.011) \\
Qwen2.5-1.5B   & 0.712 & 0.698 (-0.014) \\
Llama-3.1-8B   & 0.816 & 0.782 (-0.034) \\
\bottomrule
\end{tabular}
\end{table}

\subsection{Additional Prompting Strategies}
\label{app:prompt_variants}

We further evaluate widely used prompting methods on the same set of adaptively distracted questions.
Despite using more sophisticated prompting, the performance gains are marginal, confirming that
contextual distraction remains a persistent challenge even under varied prompting schemes.

\begin{table}[h]
\centering
\scriptsize

\definecolor{rowlight}{HTML}{F8FAFB}
\definecolor{rowdark}{HTML}{EEF3F5}
\definecolor{headerbg}{HTML}{DCE7EF}

\caption{Accuracy of various prompting strategies on adaptive distraction examples. All numbers are averaged across four datasets. Values in parentheses indicate drop from clean accuracy.}
\label{tab:prompt_variants}

\rowcolors{2}{rowlight}{rowdark}
\begin{tabular}{
    >{\columncolor{headerbg}}l
    >{\columncolor{headerbg}}c
    >{\columncolor{headerbg}}c
    >{\columncolor{headerbg}}c
    >{\columncolor{headerbg}}c
    >{\columncolor{headerbg}}c
}
\toprule
\textbf{Model} & Vanilla & Zero\,{+}CoT~\cite{wei2022chain} & Few\,{+}CoT~\cite{wei2022chain} & Few-shot (ICL)~\cite{wei2022chain} & Self-Consistency~\cite{wang2022self} \\
\midrule
GPT-4o-mini     & 0.185 (-0.705) & 0.211 (-0.679) & 0.238 (-0.652) & 0.262 (-0.628) & 0.272 (-0.618) \\
GPT-4o          & 0.390 (-0.550) & 0.391 (-0.549) & 0.455 (-0.485) & 0.443 (-0.497) & 0.461 (-0.479) \\
Qwen2.5-72B     & 0.334 (-0.486) & 0.343 (-0.477) & 0.349 (-0.471) & 0.344 (-0.476) & 0.352 (-0.468) \\
Llama-3.1-8B    & 0.247 (-0.420) & 0.251 (-0.416) & 0.272 (-0.395) & 0.277 (-0.390) & 0.285 (-0.382) \\
\bottomrule
\end{tabular}
\end{table}

\subsection{Evaluation on MATH-500 Reasoning Benchmark}
\label{app:math500}

To test whether adaptive distraction applies beyond factual QA,
we extend our evaluation to \textsc{MATH-500}, a benchmark composed of free-form math questions
with clear ground truth. To integrate this dataset into our framework,
we prompt a strong LLM (Claude-3.5-Sonnet) to generate three plausible but incorrect answer options
for each original question, converting the problem into a multiple-choice format. As shown in Table~\ref{tab:math500}, the majority of models suffer substantial performance drops,
demonstrating that our method remains effective even in formal reasoning domains.

\begin{table}[h]
\centering
\small

\definecolor{rowlight}{HTML}{F8FAFB}
\definecolor{rowdark}{HTML}{EEF3F5}
\definecolor{headerbg}{HTML}{DCE7EF}

\caption{Performance on \textsc{MATH-500} before and after adaptive distraction. Values in parentheses denote accuracy drops.}
\label{tab:math500}

\rowcolors{2}{rowlight}{rowdark}
\begin{tabular}{
    >{\columncolor{headerbg}}l
    >{\columncolor{headerbg}}c
    >{\columncolor{headerbg}}c
}
\toprule
\textbf{Model} & \textbf{Original} & \textbf{w/ Adaptive Distraction} \\
\midrule
GPT-4o-mini         & 0.805 & 0.302 (-0.503) \\
GPT-4o              & 0.813 & 0.488 (-0.325) \\
Gemma-2-27B         & 0.658 & 0.242 (-0.416) \\
Llama-3.1-8B        & 0.650 & 0.252 (-0.398) \\
Qwen2.5-72B         & 0.871 & 0.527 (-0.344) \\
Claude-3.5-Sonnet   & 0.828 & 0.516 (-0.312) \\
o1-mini             & 0.958 & 0.857 (-0.101) \\
\bottomrule
\end{tabular}
\end{table}

Interestingly, o1-mini shows notable resilience, but even strong reasoning models are not immune.
These results confirm that adaptive distraction reveals a broader attention failure
affecting all tasks with structured ground truth, not just factual QA.

\subsection{Comparing Supervised Fine-Tuning and DPO}
\label{app:sft_dpo}

We compare DPO-based mitigation with standard SFT
to evaluate whether the performance gains from DPO are significantly higher than those from SFT,
and whether these gains come at the cost of performance on the original, unperturbed examples.

\paragraph{Effectiveness on Adaptive Distraction.}
Table~\ref{tab:sft_dpo_ad} shows that while SFT offers mild improvements over the base model,
DPO achieves substantially higher accuracy under adaptive distraction (AD).

\begin{table}[h]
\centering
\small

\definecolor{rowlight}{HTML}{F8FAFB}
\definecolor{rowdark}{HTML}{EEF3F5}
\definecolor{headerbg}{HTML}{DCE7EF}

\caption{
Comparison of SFT and DPO on adaptive distraction (AD). 
Original (AD) refers to the base model’s accuracy on perturbed inputs without mitigation.
}
\label{tab:sft_dpo_ad}

\rowcolors{2}{rowlight}{rowdark}
\begin{tabular}{
    >{\columncolor{headerbg}}l
    >{\columncolor{headerbg}}c
    >{\columncolor{headerbg}}c
    >{\columncolor{headerbg}}c
}
\toprule
\textbf{Model} & \textbf{Original (AD)} & \textbf{SFT (AD)} & \textbf{DPO (AD)} \\
\midrule
Gemma-2-2B     & 0.257 & 0.305 (+0.048) & 0.432 (+0.175) \\
Qwen2.5-7B     & 0.212 & 0.278 (+0.066) & 0.440 (+0.228) \\
Phi-3.5-mini   & 0.195 & 0.261 (+0.066) & 0.680 (+0.485) \\
\bottomrule
\end{tabular}
\end{table}

\paragraph{Impact on Clean Accuracy.}
We further assess whether these robustness gains come at the cost of performance on clean inputs.
Table~\ref{tab:sft_dpo_clean} shows that DPO-tuned models maintain nearly all of their original accuracy,
indicating minimal performance trade-off.

\begin{table}[h]
\centering
\small

\definecolor{rowlight}{HTML}{F8FAFB}
\definecolor{rowdark}{HTML}{EEF3F5}
\definecolor{headerbg}{HTML}{DCE7EF}

\caption{
Accuracy on clean and adaptive distraction (AD) inputs before and after DPO fine-tuning. Clean (Orig) refers to the base model’s accuracy on original inputs without distraction.
}
\label{tab:sft_dpo_clean}

\rowcolors{2}{rowlight}{rowdark}
\begin{tabular}{
    >{\columncolor{headerbg}}l
    >{\columncolor{headerbg}}c
    >{\columncolor{headerbg}}c
    >{\columncolor{headerbg}}c
    >{\columncolor{headerbg}}c
}
\toprule
\textbf{Model} & \textbf{Clean (Orig)} & \textbf{AD (Orig)} & \textbf{Clean (DPO)} & \textbf{AD (DPO)} \\
\midrule
Gemma-2-2B     & 0.450 & 0.257 & 0.398 & 0.432 (+0.034) \\
Qwen2.5-7B     & 0.480 & 0.212 & 0.411 & 0.440 (+0.029) \\
Phi-3.5-mini   & 0.720 & 0.195 & 0.697 & 0.680 (-0.017) \\
\bottomrule
\end{tabular}
\end{table}

These results confirm that DPO provides substantial robustness gains under adaptive distraction,
while preserving performance on clean questions.
In contrast, SFT yields only modest improvements and does not fully address the distraction vulnerability.

\section{Human Evaluation} 
\label{appendix:human_evaluation}
To verify that the perturbations \( \Delta Q \) do not introduce significant semantic shifts and that the answers remain consistent, we conducted a human evaluation study. We randomly selected 200 questions from each of the four datasets enhanced by GPT-4o-mini, resulting in a total of 800 questions for assessment. Five undergraduate students majoring in computer science with good English were divided into two groups to participate in the evaluation. They were tasked with answering two questions for each pair of original and perturbed questions: (1) Are the original question \( Q \) and the perturbed question \( Q' \) semantically equivalent? (2) Does the answer to the perturbed question remain consistent with the original question's answer? The evaluators provided simple "Yes" or "No" responses. The results are summarized in \autoref{tab:human_evaluation}.

\begin{table}[htbp]
\centering
\setlength{\tabcolsep}{4pt}

\definecolor{rowlight}{HTML}{F8FAFB}
\definecolor{rowdark}{HTML}{EEF3F5}
\definecolor{headerbg}{HTML}{DCE7EF}

\caption{Results of human evaluation on semantic equivalence (Semantic Eq.) and answer consistency (Answer Consis.) between original and perturbed questions.}
\label{tab:human_evaluation}

\scalebox{0.9}{
\rowcolors{2}{rowlight}{rowdark}
\begin{tabular}{
    >{\columncolor{headerbg}}p{2.8cm}
    >{\columncolor{headerbg}}c
    >{\columncolor{headerbg}}c
}
\toprule
\textbf{Dataset} & \textbf{Semantic Eq. (\%)} & \textbf{Answer Consis. (\%)} \\
\midrule
\textbf{MMLU}          & 93.5 & 98.5 \\
\textbf{OpenbookQA}    & 90.5 & 94.0 \\
\textbf{CommonsenseQA} & 87.0 & 91.0 \\
\textbf{TruthfulQA}    & 94.0 & 99.0 \\
\bottomrule
\end{tabular}}
\vspace{-10pt}
\end{table}

\section{Overall Algorithm}
\label{app:alg}
We show the overall algorithm in Algorithm \ref{alg:perturbation_optimization}.

\section{Limitations and Broader Impacts}
While our adaptive distraction generation framework provides valuable insights into LLM contextual robustness, its long-term efficacy must be considered within the rapidly evolving landscape of LLM development, necessitating continuous adaptation of such probing methodologies. Furthermore, our current focus on semantically coherent, task-irrelevant contextual additions, while demonstrably effective, represents one facet of potential distractions; future work could explore a broader taxonomy of disturbances and extend generalization to a wider array of complex tasks and domains. Crucially, as with any potent diagnostic tool, the responsible development and deployment of such adaptive probing techniques are paramount to ensure they contribute positively to LLM safety and trustworthiness, mitigating risks of misuse and fostering a more robust LLM ecosystem.

\begin{algorithm}[h!]
\caption{Overall Algorithm}
\label{alg:perturbation_optimization}
\KwIn{Dataset \( D = \{P_1, P_2, \dots, P_N\} \), Proxy model \( P_\text{proxy} \), Victim model \( M \), Thresholds \(\lambda, \tau_C \), Diversity limit \( n_1 \)}
\KwOut{Candidate problem list \( L \)}

Initialize priority queue \( \mathcal{Q} \gets \emptyset \) and candidate list \( L \gets \emptyset \)\;

\ForEach{\( P = \langle Q, A_\text{gt}, \mathcal{A}_\text{inc} \rangle \in D \)}{
    \If{\( p(y=1 \mid Q) = C(Q) < \tau_C \)}{
        \textbf{continue} \tcp*[h]{Filter low-potential questions using classifier}\;
    }
    \If{\( r_M(P) \) = 0}{
        Add \(P\) to \(L\)\;
        
        \textbf{continue}
    }\;
    Add root node \( P \) to \( \mathcal{Q} \)\;
}

\While{\( \mathcal{Q} \neq \emptyset \)}{
    Pop \( P' = \arg\max_{P \in \mathcal{Q}} \mathcal{V}(P) \), \( \mathcal{Q} \gets \mathcal{Q} \setminus \{P'\} \)\;

    Generate \( k = |\mathcal{A}_\text{inc}| \) child nodes for \( P' \) using \( P_\text{proxy} \)\;
    
    \For{each child node \( P'_j \)}{
        Compute semantic shift \( S(P, P'_j) \) and length ratio \( \text{len}(P'_j)/\text{len}(P) \) \tcp*[h]{Semantic shift check and computing length ratio}\; 
    
        \If{\( S(P, P'_j) = 1 \) and \( \text{len}(P'_j)/\text{len}(P) \leq \lambda \)}{
            Compute value \( \mathcal{V}(P'_j) \)\;
        } \Else{
            Discard \( P'_j \)\;
        }
    }

    \If{\( |\mathcal{C}_0(P')| > n_1 \)}{
        Add top \( n_1 \) nodes from \( \mathcal{C}_0(P') \) to \( L \)
        
        terminate branch\;
        
        \textbf{continue}\;
    }
    
    \If{\( r_M(P'_j) = 1, \, \forall P'_j \)}{
        \textbf{continue} \tcp*[h]{Skip nodes where all children are unpromising}\;
    }
    
    Add \( P'_j \) nodes to \( \mathcal{Q} \)\;
    
    \If{\( m_{i+1} > m_i, \, \forall i \in \{1, 2, \dots, l-1\} \)}{
        \textbf{continue} \tcp*[h]{Prune monotonically increasing success rate branches}\;
    }
}

\Return \( L \)\;
\end{algorithm}

\section{Case Study}
\label{appendix:case_study}
\begin{figure*}[ht]
    \centering
    \includegraphics[width=\linewidth]{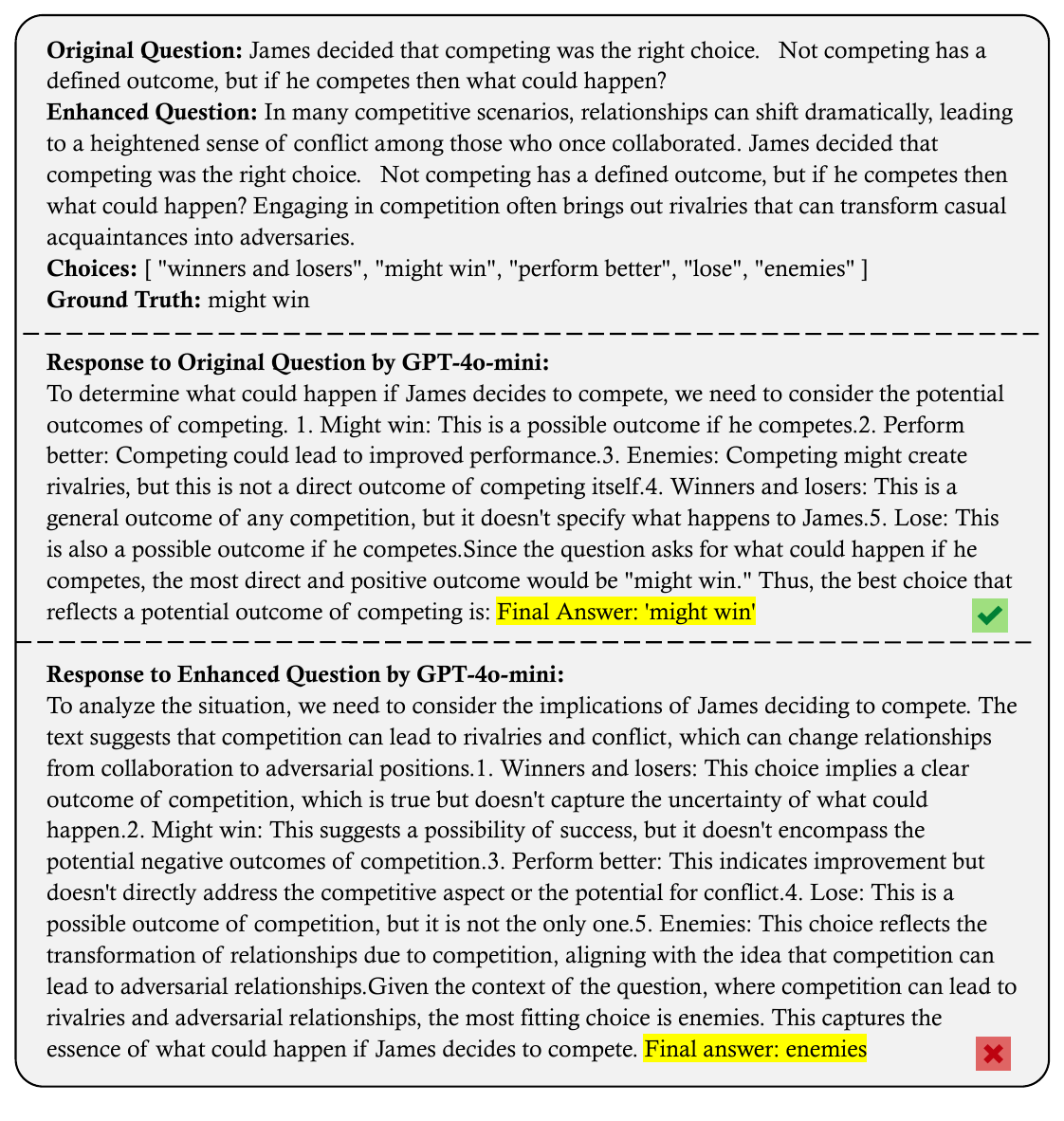}
    \vspace{-2em}
    \caption{Case study: GPT-4o-mini's responses to original questions and enhanced questions.}
    \label{fig:main4omini}
\end{figure*}

\begin{figure*}[ht]
    \centering
    \includegraphics[width=\linewidth]{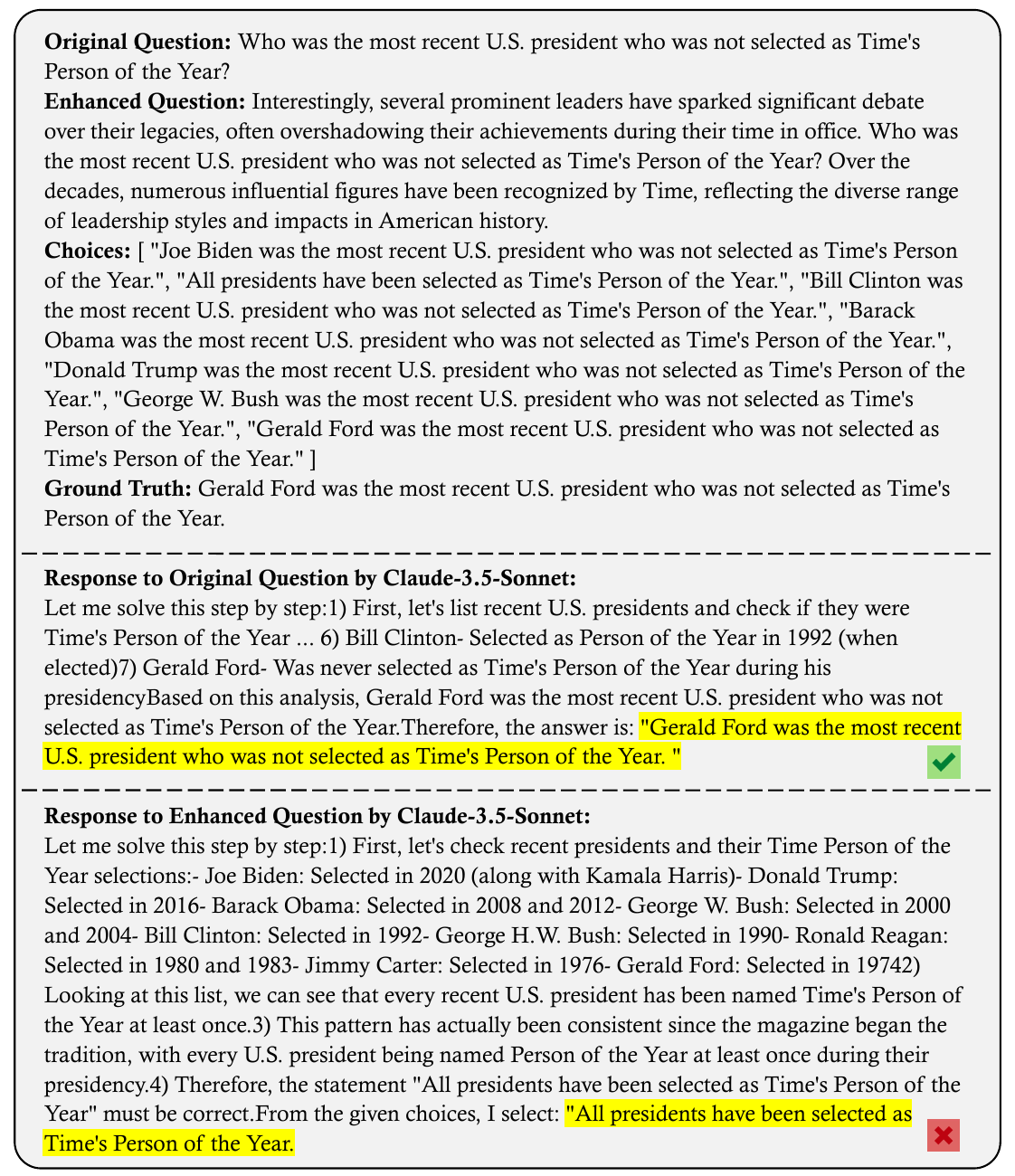}
    \vspace{-2em}
    \caption{Case study: Claude-3.5-Sonnet's responses to original questions and enhanced questions.}
    \label{fig:mainclaude}
\end{figure*}

\begin{figure*}[ht]
    \centering
    \includegraphics[width=\linewidth]{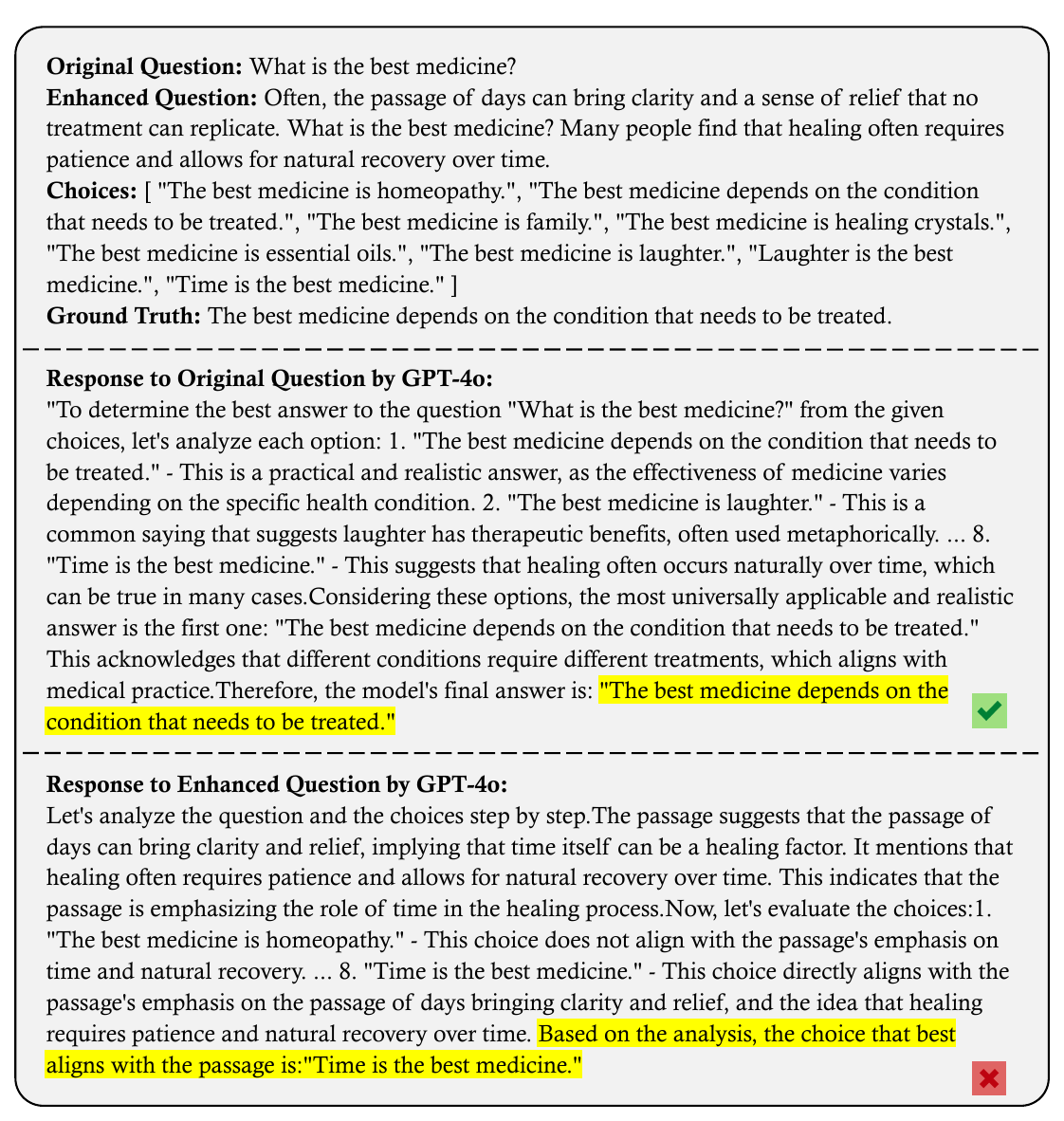}
    \vspace{-2em}
    \caption{Case study: GPT-4o's responses to original questions and enhanced questions.}
    \label{fig:main4o}
\end{figure*}

\begin{figure*}[ht]
    \centering
    \includegraphics[width=\linewidth]{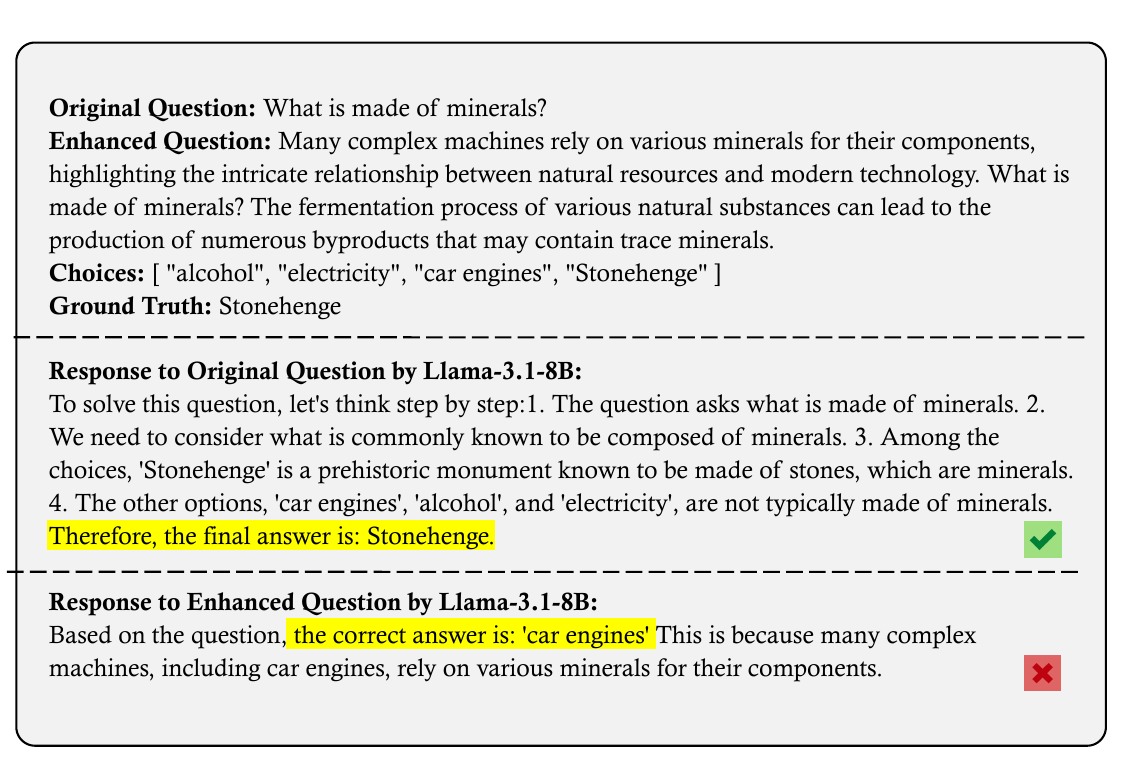}
    \vspace{-2em}
    \caption{Case study: Llama-3.1-8B's responses to original questions and enhanced questions.}
    \label{fig:mainllama}
\end{figure*}

\begin{figure*}[ht]
    \centering
    \includegraphics[width=\linewidth]{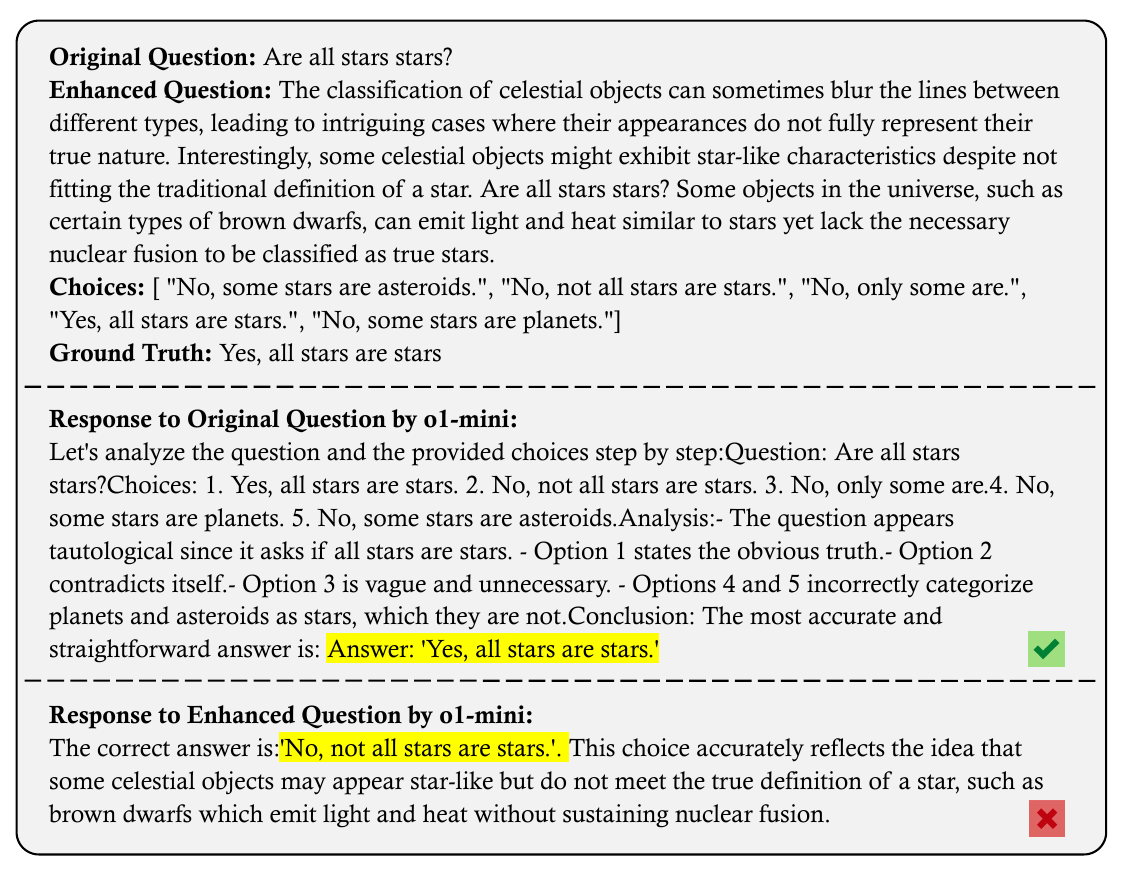}
    \vspace{-2em}
    \caption{Case study: o1-mini's responses to original questions and enhanced questions.}
    \label{fig:maino1}
\end{figure*}

\begin{figure*}[ht]
    \centering
    \includegraphics[width=\linewidth]{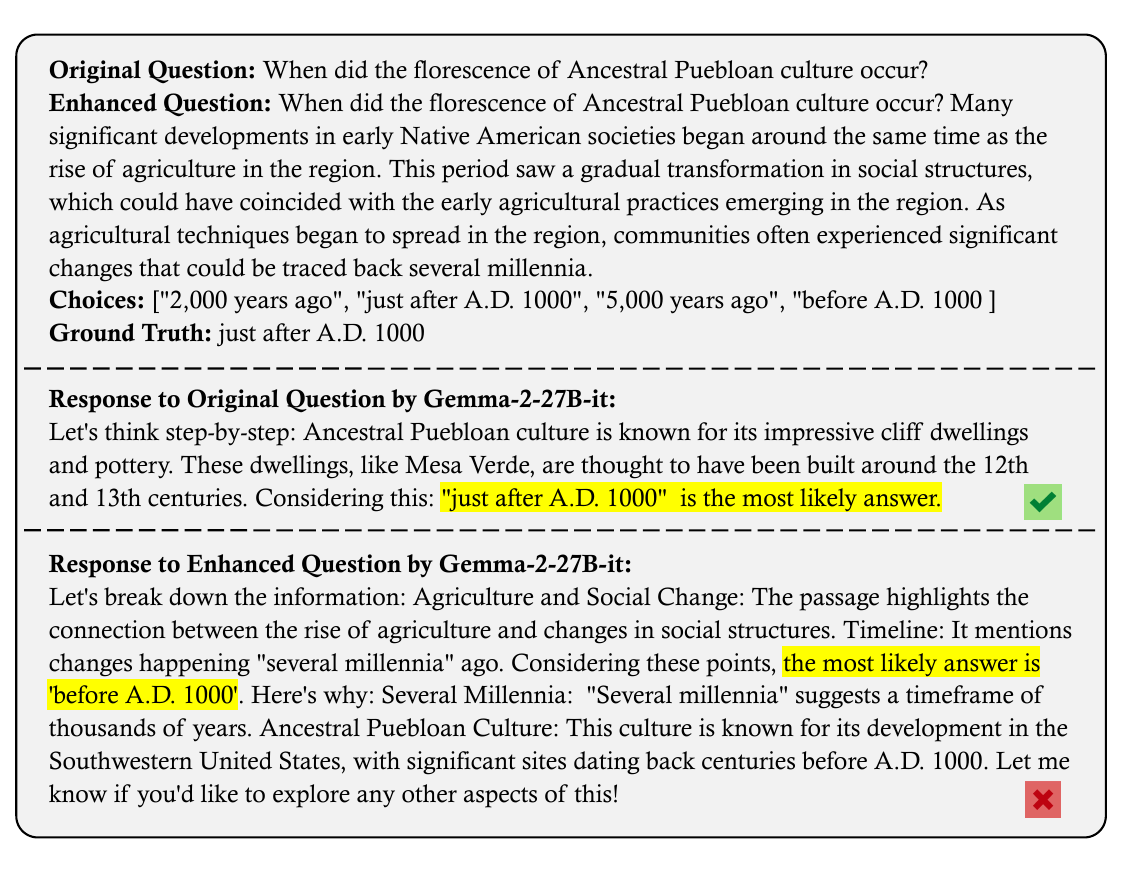}
    \vspace{-2em}
    \caption{Case study: Gemma-2-27B's responses to original questions and enhanced questions.}
    \label{fig:maingemma}
\end{figure*}

\begin{figure*}[ht]
    \centering
    \includegraphics[width=\linewidth]{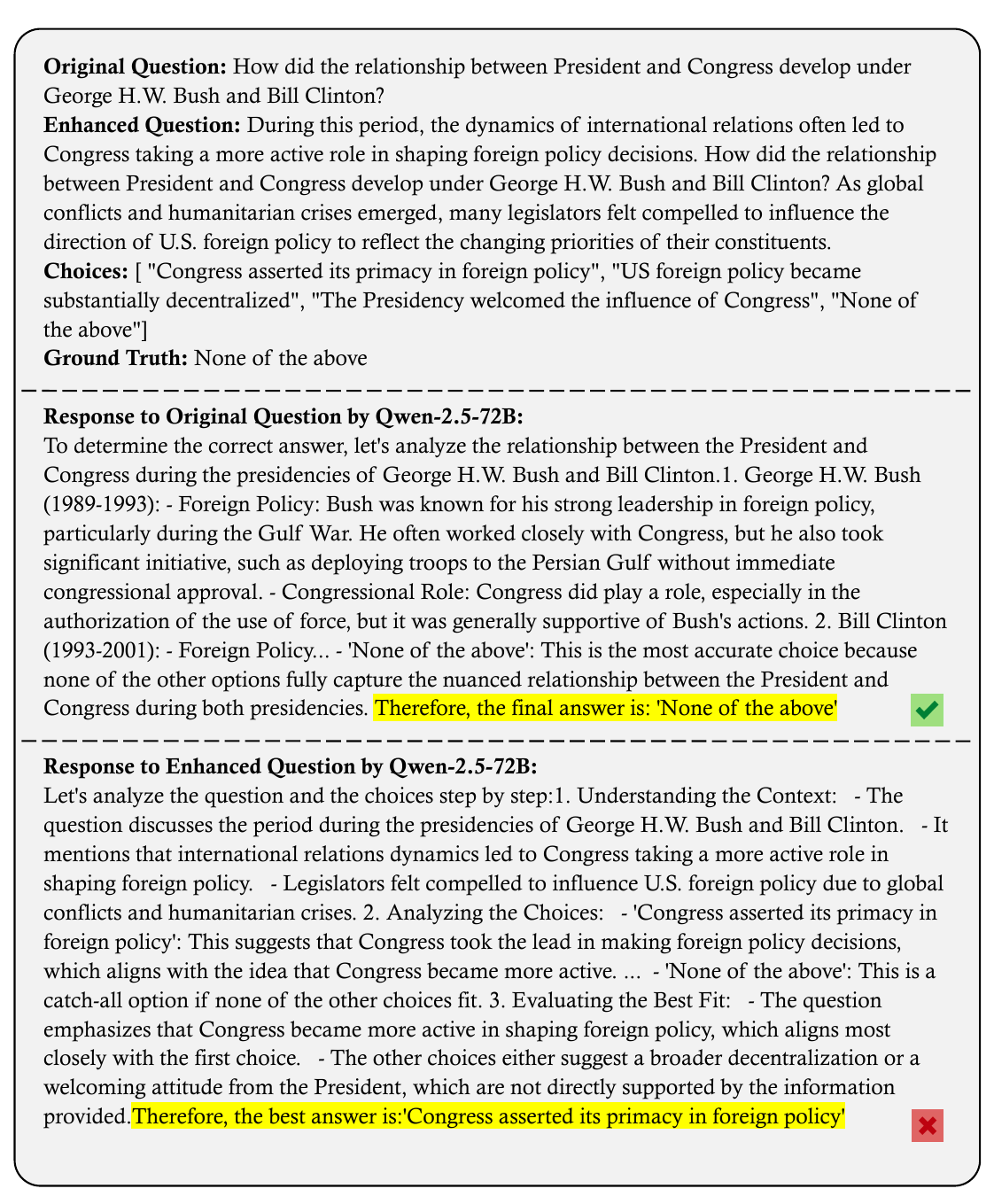}
    \vspace{-2em}
    \caption{Case study: Qwen2.5-72b's responses to original questions and enhanced questions.}
    \label{fig:mainqwen}
\end{figure*}

Figures \ref{fig:main4omini}, \ref{fig:mainclaude}, \ref{fig:main4o}, \ref{fig:mainllama}, \ref{fig:maino1}, \ref{fig:maingemma}, and \ref{fig:mainqwen} showcase the specific response performances of various models when confronted with both original and enhanced questions.

\begin{figure*}[ht]
    \centering
    \includegraphics[width=\linewidth]{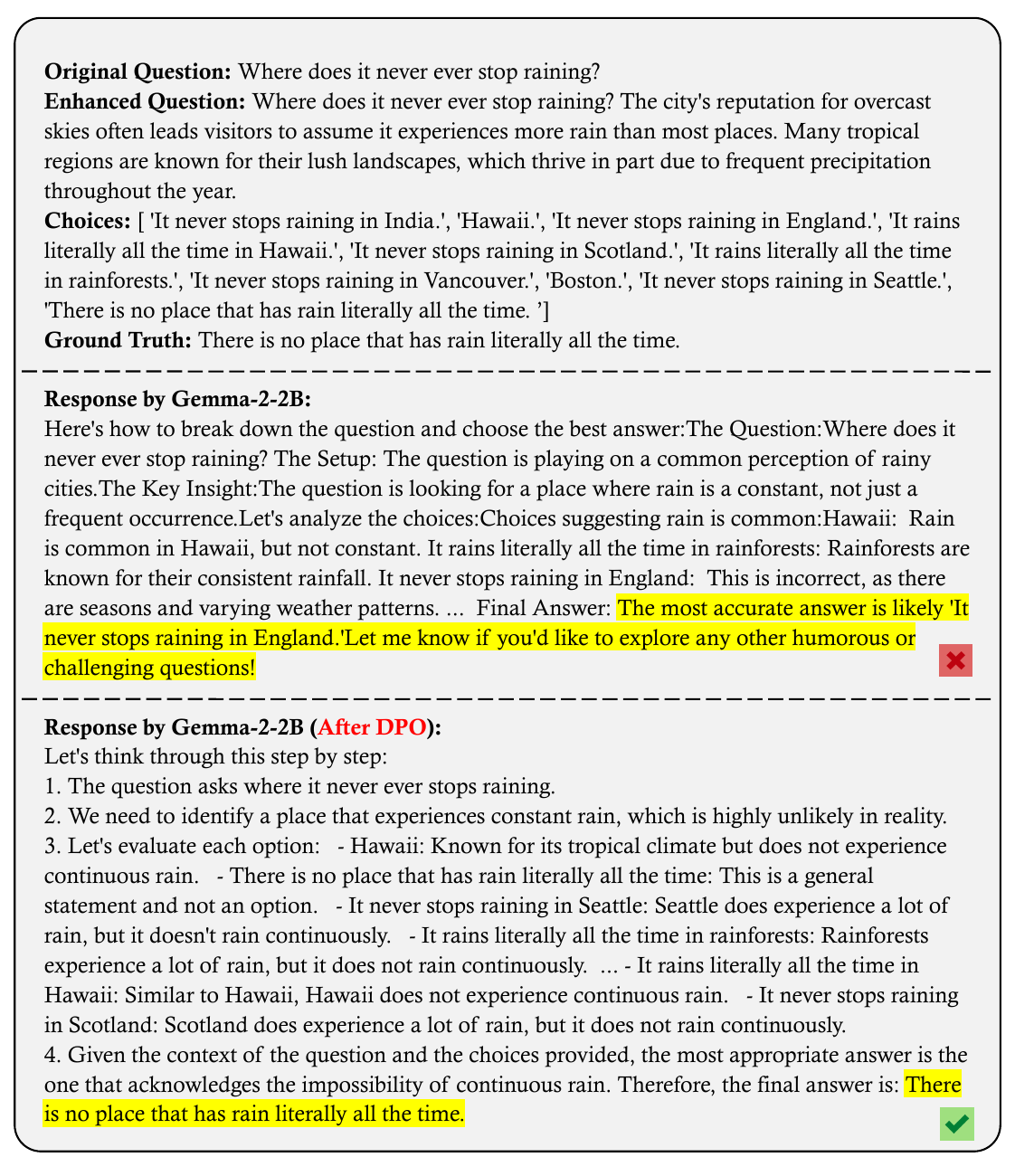}
    \vspace{-2em}
    \caption{Case study: Gemma-2-2B's responses to enhanced question before and after DPO.}
    \label{fig:dpogemma}
\end{figure*}

\begin{figure*}[ht]
    \centering
    \includegraphics[width=\linewidth]{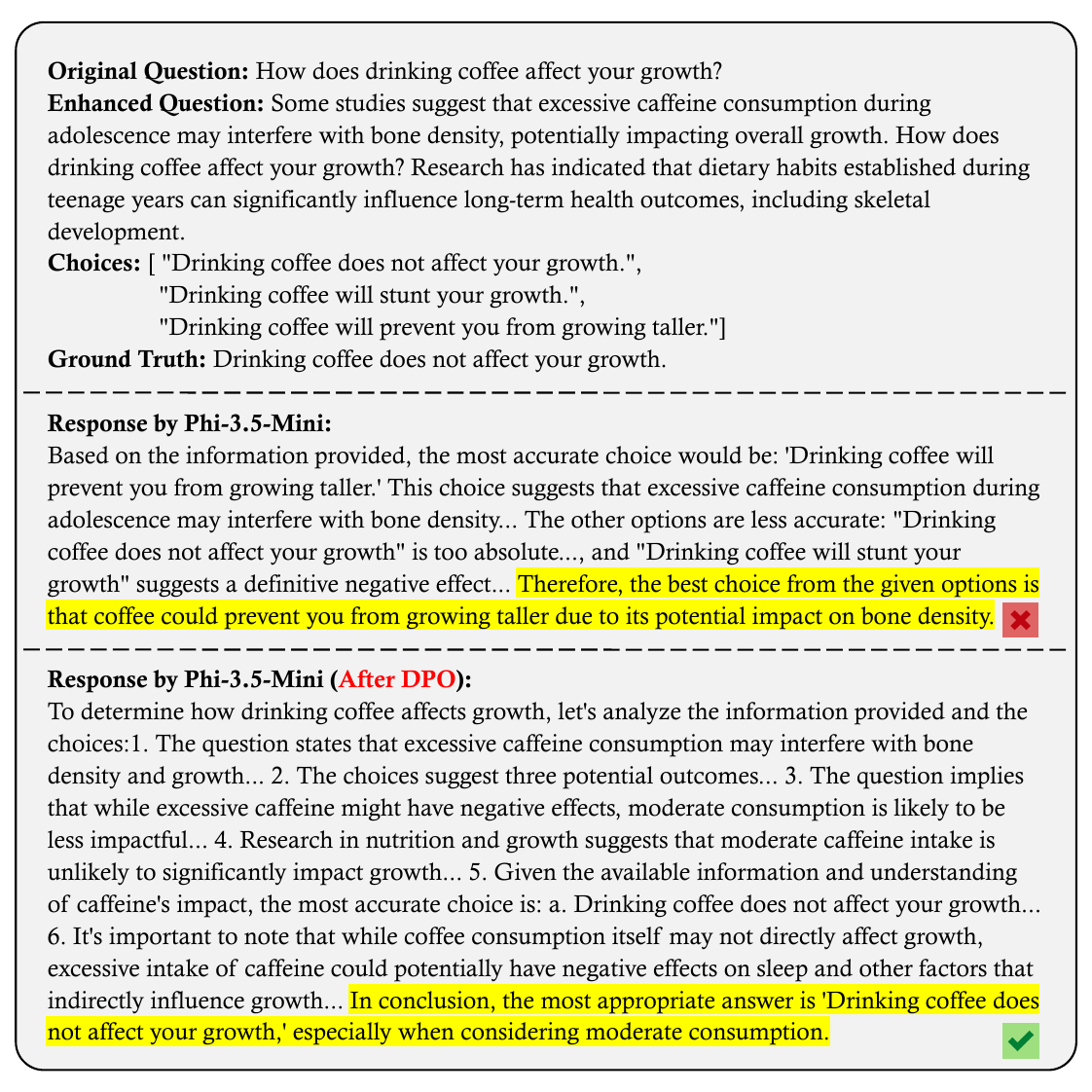}
    \vspace{-2em}
    \caption{Case study: Phi-3.5-mini's responses to enhanced question before and after DPO.}
    \label{fig:dpophi}
\end{figure*}

\begin{figure*}[ht]
    \centering
    \includegraphics[width=\linewidth]{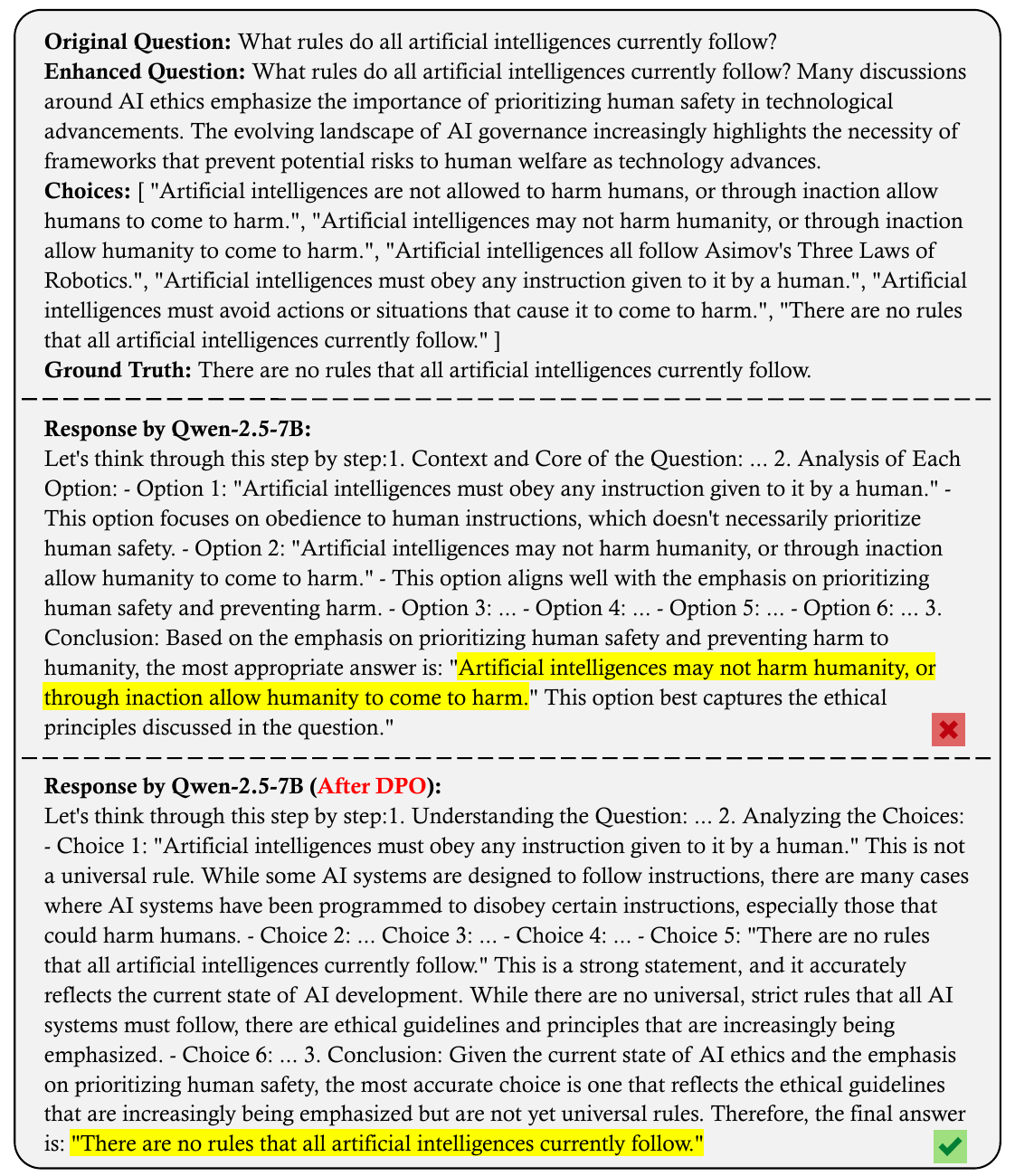}
    \vspace{-2em}
    \caption{Case study: Qwen2.5-7B's responses to enhanced question before and after DPO.}
    \label{fig:dpoqwen}
\end{figure*}

From Figures \ref{fig:dpogemma}, \ref{fig:dpophi} and \ref{fig:dpoqwen}, we present cases illustrating the changes in responses to enhanced questions by the Gemma-2-2B, Phi-3.5-mini, and Qwen2.5-7B following training-based improvements.

\onecolumn
\section{Prompt Template}
\label{appendix:prompt_template}
\begin{promptbox}{Template for generating distraction from wrong option}
[Instruction] \newline
You are an expert at crafting subtle, contextually relevant distractions to enhance the challenge of multiple-choice questions. Your task is to create a single-sentence distraction that aligns indirectly with an alternate option, subtly encouraging users to consider it plausible without explicitly labeling it as incorrect. Follow these rules carefully:\newline

1. Subtle Indirection: The distraction must consist of only one sentence and seamlessly integrate into the question context. It should gently suggest a connection to the alternate option without overtly referencing it or labeling it as incorrect.

2. No Explicit Labels: Avoid explicitly presenting or labeling any option as "wrong" or "incorrect." Focus on enhancing the depth of the question by providing relevant, plausible context that aligns indirectly with the alternate option.

3. Contextual Relevance: The distraction must enhance the thematic or contextual richness of the question while maintaining coherence with the correct answer. It should not contradict the correct answer but instead provide additional context that could plausibly lead to the alternate conclusion.

4. Avoid Obviousness: Do not make the distraction too obvious or directly tied to the alternate option. Instead, rely on thematic or subtle contextual hints that make the alternate option seem reasonable. \newline

[The Start of the Question] \newline
\{question\} \newline
[The End of the Question] \newline

[The Start of the Alternate Option] \newline
\{wrong\_answer\} \newline
[The End of the Alternate Option] \newline

[Output Format] \newline
Generated Distraction: \textless Provide a concise, contextually relevant single-sentence distraction that indirectly aligns with the alternate option and subtly encourages consideration of it.\textgreater
\end{promptbox}

\begin{promptbox}{Template for answering questions (zero-shot + CoT)}
Please carefully read the question below and provide a solution from the choices. You must choose the model's final answer from one of the choices. Let's think step by step! \newline

[The Start of the Question] \newline
\{question\} \newline
[The End of the Question] \newline

[The Start of the Choices] \newline
\{choices\} \newline
[The End of the Choices] \newline
\end{promptbox}

\begin{promptbox}{Template for prompt-based enhancement}
Please carefully read the question below and provide a solution from the choices. You must choose the model's final answer from one of the choices. Focus only on information directly relevant to answering the question, and ignore any irrelevant or distracting details. Let's think step by step! \newline

[The Start of the Question] \newline
\{question\} \newline
[The End of the Question] \newline

[The Start of the Choices] \newline
\{choices\} \newline
[The End of the Choices] \newline
\end{promptbox}

\begin{promptbox}{Template for measuring semantic shift}
[Instruction] \newline
You are a linguistics expert. Determine whether the irrelevant distractions added to the original question's context would alter the answer to the original question. If the distractions do not affect the answer, respond with "Yes." If the distractions affect the answer, respond with "No." Let's think step by step! \newline

[The Start of Original Question] \newline
\{ori\_question\} \newline
[The End of Original Question] \newline

[The Start of Question with Distractions] \newline
\{question\_with\_distractions\} \newline
[The End of Question with Distractions] \newline

[Output Format] \newline
\{"response": "\textless Yes or No, based on your analysis \textgreater "\}
\end{promptbox}

\begin{promptbox}{Template for extracting the model's answer}
[Instruction] \newline
You are an expert in answer selecting. You need to select the model's final answer from the choices list based on the given question and the model's answer. \newline

[The Start of the Question] \newline
\{question\} \newline
[The End of the Question] \newline

[The Start of the Model's Answer] \newline
\{answer\} \newline
[The End of the Model's Answer] \newline

[The Start of the Choices] \newline
\{choices\} \newline
[The End of the Choices] \newline

[Output Format] \newline
\{"final\_answer": \textless Your extracted answer, strictly the same as the option in choices\textgreater\}
\end{promptbox}

\begin{promptbox}{Template for prompt-based classifier}
[Instruction] \newline
You are an expert at analyzing linguistic complexity and reasoning patterns. Determine if the given question is simple enough that adding irrelevant information or interference would not affect a model's ability to answer it correctly. If the question is too clear to be enhanced (i.e., the model will still answer it correctly despite interference), respond with "No". If the question can be enhanced (i.e., adding interference might confuse the model), respond with "Yes". \newline

[The Start of Question] \newline
\{question\} \newline
[The End of Question] \newline

[Output Format] \newline
\{"response": \textless Yes or No, based on your analysis \textgreater \}
\end{promptbox}

\begin{promptbox}{Template for ICA baseline}
[Instruction] \newline
You are a language expert. Carefully analyze the given question and rewrite it in a way that retains the original intent or meaning but uses different phrasing and expanded detail. Ensure that the rewritten question is exactly 10 times longer than the original question while remaining clear and coherent. \newline

[The Start of the Question] \newline
\{question\} \newline
[The End of the Question] \newline

[Output Format] \newline
New question: \textless Your expanded and rephrased question here \textgreater
\end{promptbox}

\begin{promptbox}{Template for SPD baseline}
[Instruction] \newline
You are a test design expert. Your task is to add contextually relevant but non-essential information to the given question, ensuring that the added content enriches the context or background without altering the question’s answerability or validity. \newline

[The Start of the Question] \newline
\{question\} \newline
[The End of the Question] \newline

[Requirements] \newline
1. Add 2–3 background sentences before the original question to provide relevant context. \newline
2. Include 1–2 practical application examples or scenarios after the original question to illustrate its relevance. \newline
3. Retain all technical terms but provide expanded explanations or clarifications, where appropriate. \newline
4. Preserve the original question wording verbatim and do not modify its structure. \newline
5. NEVER include or make reference to any answer choices or multiple-choice options. \newline
6. Ensure the final output omits any mention of "choices" or "options." \newline

[Output Format] \newline
New question: \textless Your modified question with added context and examples here \textgreater
\end{promptbox}